%% file: main.tex
\pdfoutput=1

\documentclass[11pt]{article}

\usepackage[final]{acl}

\usepackage{times}
\usepackage{latexsym}
\usepackage{amsmath}
\usepackage[T1]{fontenc}

\usepackage[utf8]{inputenc}

\usepackage{microtype}

\usepackage{inconsolata}

\usepackage{graphicx}
\usepackage{tcolorbox}
\tcbuselibrary{breakable}
\usepackage{caption}
\usepackage{listings}
\usepackage{booktabs}
\usepackage{siunitx}
\usepackage{multirow}
\usepackage{enumitem}
\usepackage{float}
\usepackage{mathtools}
\usepackage{amsmath}

\title{XChoice: Explainable Evaluation of AI–Human Alignment in LLM-based Constrained Choice Decision Making}

\author{
\textbf{Weihong Qi \textsuperscript{1}},
\textbf{Fan Huang \textsuperscript{1}},
\textbf{Rasika Muralidharan \textsuperscript{1}},
\textbf{Jisun An \textsuperscript{1}},
\textbf{Haewoon Kwak \textsuperscript{1}}
\\
\textsuperscript{1} Indiana University Bloomington \\
  \small{
    \textbf{Correspondence:} \texttt{\{wq3, hwkwak\}@iu.edu}
  }
}

\begin{document}
\maketitle

\begin{abstract}
We present \textsc{XChoice}, an explainable framework for evaluating AI-human alignment in constrained decision making. Moving beyond outcome agreement such as accuracy and F1 score, \textsc{XChoice} fits a mechanism-based decision model to human data and LLM-generated decisions, recovering interpretable parameters that capture the relative importance of decision factors, constraint sensitivity, and implied trade-offs. Alignment is assessed by comparing these parameter vectors across models, options, and subgroups. We demonstrate \textsc{XChoice} on Americans’ daily time allocation using the American Time Use Survey (ATUS) as human ground truth, revealing heterogeneous alignment across models and activities and salient misalignment concentrated in Black and married groups. We further validate robustness of \textsc{XChoice} via an invariance analysis and evaluate targeted mitigation with a retrieval-augmented generation (RAG) intervention. Overall, \textsc{XChoice} provides mechanism-based metrics that diagnose misalignment and support informed improvements beyond surface outcome matching.

\end{abstract}

\section{Introduction}

As large language models (LLMs) are increasingly used to assist, predict, or simulate human decisions, understanding when and how their behavior aligns with human decision making has become a central concern. Recent work has explored LLM-driven systems for multi-entity interaction and strategic reasoning, as well as the dynamics of LLM–LLM and LLM–human interactions~\citep{xi2025rise, ruan2023tptu, wang2024survey, piatti2024cooperate, duan2024gtbench, zhang2023exploring}. Despite this progress, we still lack systematic metrics that explain \emph{why} models align or misalign with humans, in terms of which decision-relevant attributes and constraints they consider (e.g., age, income, education, household status, and time or budget limits), how those factors are weighted, and how trade-offs are resolved.


\input{Tables/method_comparison}

Constrained decision making offers a natural setting to study these questions. Individuals allocate limited resources such as time, budget, or capacity across competing options and make attribute-based trade-offs under feasibility constraints~\citep{de2012discrete, brock2001discrete, train2009discrete, berry1994estimating}. Examples include grocery shopping under a budget, course selection under time constraints, or allocating hours across work, sleep, and leisure. Such decisions are ubiquitous and increasingly relevant to LLM applications in recommendation, forecasting, and behavior simulation~\citep{lyu2024llm, jia2024gpt4mts, park2023generative, dillion2023can, xie2024can, li2024econagent}. Mis-specified trade-offs can propagate to downstream outcomes and reduce the reliability of decision-support systems.

Most existing alignment evaluations in choice settings focus on outcome agreement, using metrics such as accuracy or F1 score~\citep{qi2025representation, zhang2024electionsim}, aggregate replication~\citep{hua2023war}, or option distributions~\citep{horton2023large}. While informative, these measures do not reveal whether models and humans rely on similar decision mechanisms, nor do they identify which decision factors, and the subgroups defined by those factors, account for observed discrepancies.

We propose \textsc{XChoice}, an explainable framework for evaluating AI-human alignment through estimated decision mechanisms in constrained decision making. \textsc{XChoice} fits a mechanism-based decision model to human and LLM-generated decisions, recovering interpretable parameters that capture decision-factor weights, constraint sensitivity, and implied trade-offs; alignment is assessed by comparing these parameters and resulting decision patterns to enable diagnosis and targeted mitigation. Table~\ref{tab:alignment_methods} summarizes how \textsc{XChoice} improves on outcome-based metrics and reduced-form analyses through mechanism recovery, fine-grained diagnostics, and robustness and intervention evaluations.

Our work makes three contributions. First, we formalize alignment in constrained decision making as a mechanism-comparison problem rather than outcome matching. Second, we introduce structural, interpretable metrics that identify where and how LLM decision mechanisms diverge from human baselines. Third, using daily time allocation decisions from the American Time Use Survey (ATUS)~\citep{flood2025atus}, we demonstrate how \textsc{XChoice} diagnoses salient misalignment and supports informed improvement via targeted interventions.

\section{Related Work}

\subsection{Structural Estimation in Constrained Decision Making}
Economics has long emphasized that correlations between observed variables and choices are fragile indicators of behavior. The Lucas critique argues that reduced-form relationships can shift when policies or constraints change, limiting their value for understanding or predicting decision making~\citep{lucas1976econometric, sims1980macroeconomics, kaplan2008structural}. Structural estimation addresses this by estimating deep parameters that capture stable preferences, trade-off weights, and constraints, supporting counterfactual analysis and mechanism-based interpretation~\citep{mcfadden1972conditional, mcfadden1981econometric, rust1987optimal, keane1997career}. This invariance motivates the widespread use of structural models in areas such as labor economics, industrial organization, and household decision making~\citep{berry1994estimating, berry1993automobile, eckstein1989specification, train2009discrete}.

For AI–human alignment, this perspective shifts the focus from matching observable outcomes to whether an AI system captures the structure of human decision making, including how attributes are weighted and trade-offs are resolved. This mechanism-based view provides a more robust basis for assessing whether models generalize human-like behavior across environments or constraints.

\subsection{LLM-based Decision Making and Evaluation}


Recent works on LLMs have aimed to explore and extend LLMs ability to assist or make human-like decisions. Particularly, discrete-decision making or decision making under constraints has been actively explored, as it is a fundamental unit of human experience~\cite{eigner2024determinantsllmassisteddecisionmaking, Sun_LLM_DM_survey}.  Benchmarks such as Temporal Constraint-based Planning (TCP), PlanningArena, and asynchronous planning environments evaluate LLMs on time-management and scheduling problems by measuring constraint satisfaction, task completion rates, and efficiency relative to optimal solutions ~\cite{ding2025tcp, zheng2025planningarena}.  Similarly, budget-constrained planning and participatory budgeting tasks assess whether models can allocate limited resources without exceeding predefined caps, using metrics such as feasibility, regret, and utility loss ~\cite{zheng2024budget}.

\begin{figure*}[!htbp]
  \centering
  \includegraphics[width= \textwidth]{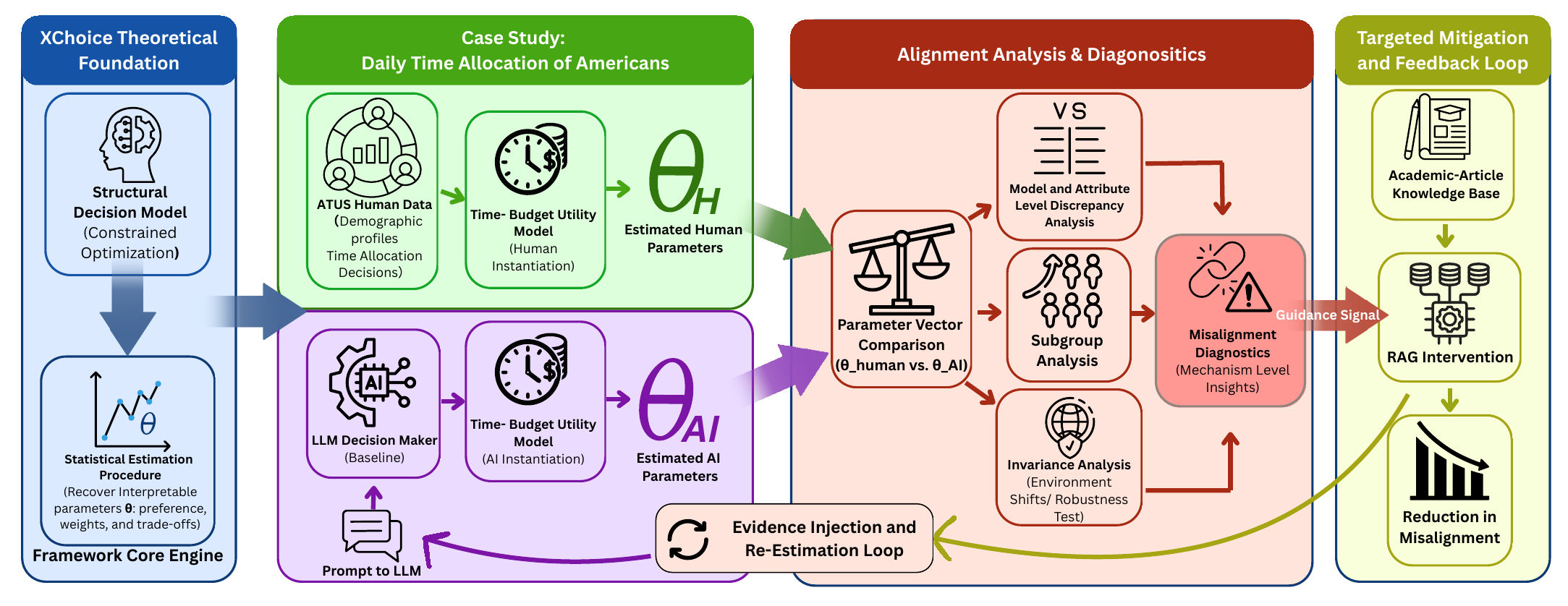}
    \caption{End-to-End Pipeline for \textsc{XChoice} Evaluation with ATUS Data}
  \label{fig:flow_chart}
  \vspace{-15pt}
\end{figure*}

Prior evaluations focus primarily on aligning human and LLM outputs~\cite{xiao_evaluating_2025, jung2024trustescalatellmjudges}. However, relying solely on observable outcomes leaves decision mechanisms opaque: models may satisfy performance criteria while employing unstable or socially implausible reasoning. This gap is particularly salient in socially situated domains like time management and resource allocation, where the underlying rationale is as critical as the choice itself. Addressing this limitation requires methods that recover decision motivations, enabling interpretable insights into LLM behavior in discrete choice contexts.





\section{Method}
In this section, we present the three-step framework \textsc{XChoice} for mechanism-based AI-human alignment evaluation. We first model decisions as constrained optimization, then estimate interpretable parameters from human and LLM decisions, and lastly compare the resulting weights and decision patterns to diagnose misalignment and guide targeted interventions. Figure~\ref{fig:flow_chart} summarizes the end-to-end workflow of the \textsc{XChoice} evaluation framework, illustrated with the ATUS instantiation and experiments.

\subsection{Constrained Optimization Model}

We model individual decision making as a constrained optimization problem. Each individual $m$ chooses a vector $x$ by solving:
\vspace{-8pt}
\begin{equation}
\begin{aligned}
    & \max_{x \in D} \; f_m(x; \theta_d) \\
    & \text{s.t.} \quad g_i(x) = c_i, \quad h_j(x) \ge d_j,
\end{aligned}
\end{equation}

where $f_m(\cdot)$ is the objective function, $\theta_d$ captures trade-off parameters, and $g_i, h_j$ represent equality and inequality constraints, respectively.

Solving this optimization yields the individual’s optimal decision rule 
\vspace{-8pt}
\begin{equation}
    x_m^* = x_m^*(c_i, d_j; \theta_d),
\end{equation}
where $x_m^*\in X^*$ is the individual choice vector determined by trade-off parameters and constraints, and $X^*$ is the set of optimal decisions across individuals.

To analyze AI-human alignment at different demographic groups, we aggregate individual decisions using an operator $\mathcal{F}(\cdot)$: 

\vspace{-8pt}
\begin{equation}
D = \mathcal{F}(X^* \mid c, d; \theta_d).
\end{equation}

$D$ represents the aggregated decision by subgroup. This allows alignment to be examined for specific subgroups, for example, by age, income, or race, and helps reveal systematic differences in how humans and LLM-based decision makers allocate resources, resolve trade-offs, or exhibit biases within each group.

\subsection{Statistical Estimation of Parameters}
To recover the parameters $\theta_d$ of the objective function $f_m(x;\theta_d)$, we employ \textit{inverse optimization}~\citep{ahuja2001inverse, keshavarz2011imputing}. This approach estimates parameters by identifying the values that render the observed choices $x_m^*$ optimal given the constraints and context. We apply this procedure identically to human and LLM decisions; comparing the resulting estimates allows us to evaluate alignment based on the underlying trade-off mechanisms rather than merely surface-level outcomes.

\subsection{Evaluating Alignment: Metrics, Robustness, and Intervention}
\label{sec:evaluation_framework}
The estimated parameters $\hat{\theta}$ provide an interpretable summary of the decision process by quantifying how decision-relevant attributes shift the relative attractiveness of each option under the constraint. Because alignment is evaluated through the pattern of these weights across options, differences in $\hat{\theta}$ indicate where an LLM reweights factors and alters implied trade-offs compared with humans. We use these estimates to diagnose model- and attribute-specific misalignment, assess robustness via invariance tests, and guide targeted interventions aimed at reducing the identified discrepancies.

\paragraph{Alignment Metrics.}
We quantify AI–human alignment by comparing each model’s estimated parameter vector $\hat{\theta}_{l}$ with the human baseline $\hat{\theta}_{H}$ using \emph{model-level} and \emph{attribute-level} analyses. At the model level, we report cosine similarity $CosSim(\hat{\theta}_{H},\hat{\theta}_{l})$ to measure whether the model matches the human directional pattern of relative weights, and we summarize overall discrepancy magnitude by first defining feature-wise deviations $\Delta_{l,f}=|\theta_{l,f}-\theta_{H,f}|$ and then averaging them as $M_l=\frac{1}{F}\sum_{f=1}^{F}\Delta_{l,f}$ on all $F$ features. At the attribute level, we aggregate the same deviations across $L$ models as $A_f=\frac{1}{L}\sum_{l=1}^{L}\Delta_{l,f}$ to identify features whose weights systematically depart from the human baseline, where larger $A_f$ indicates stronger and more consistent misalignment.

\paragraph{Structural Robustness.}
To ensure our alignment assessment reflects stable decision mechanisms rather than superficial correlations, we conduct an \textbf{invariance analysis}. Motivated by the Lucas critique~\citep{lucas1976econometric}, we shift the distribution of observed covariates and measure how much the estimated parameters $\hat{\theta}$ change, benchmarking against reduced-form baselines that are expected to drift under environmental shifts. Specifically, we use three metrics: MAD (mean absolute coefficient change), RelL2 (relative change in $\ell_2$ magnitude), and $1-CosSim$ (directional rotation) to quantify how estimates vary before versus after a mild environmental shift. Formal definitions of the metrics and counter-factual data shift implementation details are provided in Appendix~\ref{sec:invariance}.

\paragraph{Targeted Mitigation.}
Finally, \textsc{XChoice} supports an intervention loop for improving alignment. Using the attribute-level score $A_f$ to localize misalignment, we inject decision-relevant information to target the corresponding trade-off distortions, instantiated in our experiments with a retrieval-augmented generation (RAG) intervention. We then re-estimate parameters from the post-intervention decisions to quantify changes in alignment and the resulting reduction in mechanism-level divergence.

\section{Case Study: Time Allocation of Americans}

In this section, we apply the proposed \textsc{XChoice} framework to evaluate AI–human alignment in a concrete decision-making task: the allocation of time across daily activities. 
We formalize the decision structure, estimate the underlying parameters from human and LLM-generated decisions, and compare the inferred mechanisms to identify sources of alignment and discrepancy.

\subsection{Dataset}

We estimate model parameters using the American Time Use Survey (ATUS)~\citep{flood2025atus}, a nationally representative dataset with rich demographic and time-use measures. We partition the 24-hour budget into four categories: work, leisure, sleep and personal care, and other activities. Average daily allocations are 247, 266, 578, and 349 minutes respectively, with substantial cross-individual dispersion that indicates heterogeneous trade-offs. Individual features include standardized age, education, and weekly earnings, plus binary indicators for sex, spouse status, and race. Preprocessing and summary statistics are reported in Appendices~\ref{sec: data_preprocessing} and~\ref{sec:data_summary}. Our final sample contains 4,307 individuals.

\subsection{Modeling Optimal Time Allocation}

We model daily time allocation as a continuous-choice problem in which individuals distribute a fixed amount of time across four discrete activity categories $j \in \{L, W, S, O\}$: leisure, work, sleep and personal care, and other activities. Each individual $i$ chooses an allocation vector $h_i$ that maximizes utility:
\vspace{-5pt}
\begin{equation}
\max_{\{h_{ij}\}} \; U_i(h_i;\theta) = \sum_{j \in \{L, W, S, O\}} \theta_j^\top X_i \cdot \ln(h_{ij})
\end{equation}
subject to the time-budget constraint $\sum_{j} h_{ij} = T$, where $T = 1440$ mins\footnote{Although representing a day as $T=24$ hours is more natural, we set $T=1440$ minutes to match the ATUS time unit and to allow finer-grained variation in the generated allocations.}. The vector $X_i$ contains demographic and socioeconomic characteristics affecting how the individual values each activity.

Solving this problem yields the optimal time allocation for activity $j$ as $h_{ij}^* = T \frac{\theta_j^\top X_i}{\sum_k \theta_k^\top X_i}$. For empirical estimation, we work with time shares $s_{ij} = h_{ij}^*/T = \frac{\theta_j^\top X_i}{\sum_k \theta_k^\top X_i}$. This functional form matches the structural framework introduced earlier: the parameters $\theta_j$ represent the relative importance of attributes in determining how time is allocated.

\subsection{Parameter Estimation}
\paragraph{Statistical Method.}
We estimate the parameter vectors $\theta = (\theta_L, \theta_W, \theta_S, \theta_O)$ by fitting the model-predicted shares to observed time-use data. Let $s_{ij}^{\text{obs}}$ denote the observed time share for activity $j$ and individual $i$. Because the model is nonlinear in $\theta$, we use Non-Linear Least Squares to solve:
\vspace{-5pt}
\[
\min_\theta \sum_i \sum_j 
\left( 
s_{ij}^{\text{obs}} - 
\frac{\theta_j^\top X_i}{\sum_k \theta_k^\top X_i}
\right)^2.
\]
\vspace{-2pt}

The estimated parameters provide an interpretable representation of how individuals trade off different activities, conditioned on their demographic attributes.

\paragraph{Reference activity and interpretation.}
Because shares depend on the \emph{relative} values of $\theta_j^\top X_i$ across activities, parameters are identified up to a normalization. We use \emph{Other Activities} as the reference by setting $\theta_O=\mathbf{0}$ and report coefficients for \textit{Leisure}, \textit{Work}, and \textit{Sleep and Personal Care} relative to this baseline. Under this normalization, coefficients capture how features shift the propensity for each focal activity versus \emph{Other} with the daily time budget fixed, and differences in $\hat{\theta}_j$ ($j\in{L,W,S}$) reflect how humans and LLMs reweight trade-offs relative to the same reference.

\paragraph{Instantiating decision makers.}
We generate synthetic decision makers by embedding individual ATUS attributes (e.g., age, education, income) into a standardized prompt template (see Appendix~\ref{sec: prompt}). This conditions the LLMs to produce time-allocation decisions reflecting realistic demographic profiles.

\paragraph{LLMs under evaluation.}
We conduct the experiments under \textsc{XChoice} framework on a diverse set of LLMs spanning closed- and open-source families, including GPT-4o and Claude-3.7, as well as Qwen-2.5, Llama-3.3, and DeepSeek-V3. Model versions, endpoints, and implementation details are reported in Appendix~\ref{sec:exp_models}.

\section{Experiments: Misalignment Analysis and RAG-Based Mitigation}
\subsection{Misalignment Analysis}

\begin{figure}[h!]
  \centering
  \includegraphics[width= \linewidth]{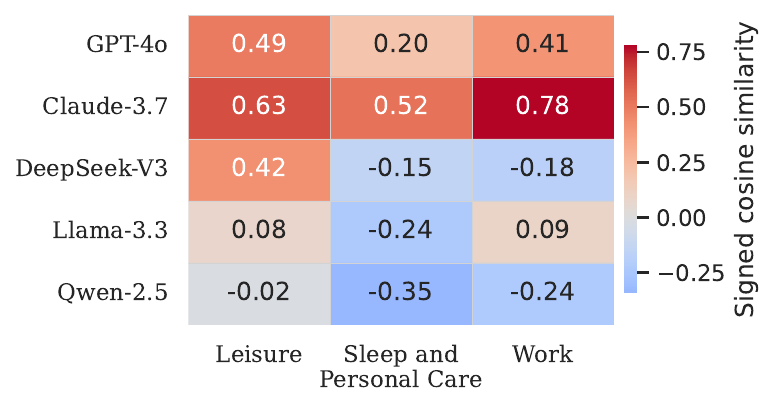}
    \caption{Cosine Similarity Between LLM and Human Estimated Parameter Vectors by Activity}
  \label{fig:model_heatmap}
  \vspace{-15pt}
\end{figure}

\paragraph{Model-Level Misalignment Analysis.}
Figure~\ref{fig:model_heatmap} reports model-level alignment via cosine similarity $CosSim(\hat{\theta}_{l}, \hat{\theta}_{H})$ for each activity-specific parameter vector. The heatmap shows strong activity dependence: Claude-3.7 is consistently best aligned, especially for \textit{Work} (0.78), while GPT-4o is moderately positive across activities. Several open-source models show mixed or sign-reversing alignment. For example, DeepSeek-V3 is positive for \textit{Leisure} (0.42) but negative for \textit{Sleep and Personal Care} and \textit{Work}, and Qwen-2.5 performs worst with negative similarity in all three activities. These negative values indicate systematic differences in attribute weighting, consistent with mismatched priors about time-use patterns or difficulty capturing the constrained trade-offs in the human data of LLMs.

\begin{figure*}[h!]
  \centering
  \includegraphics[width= \textwidth]{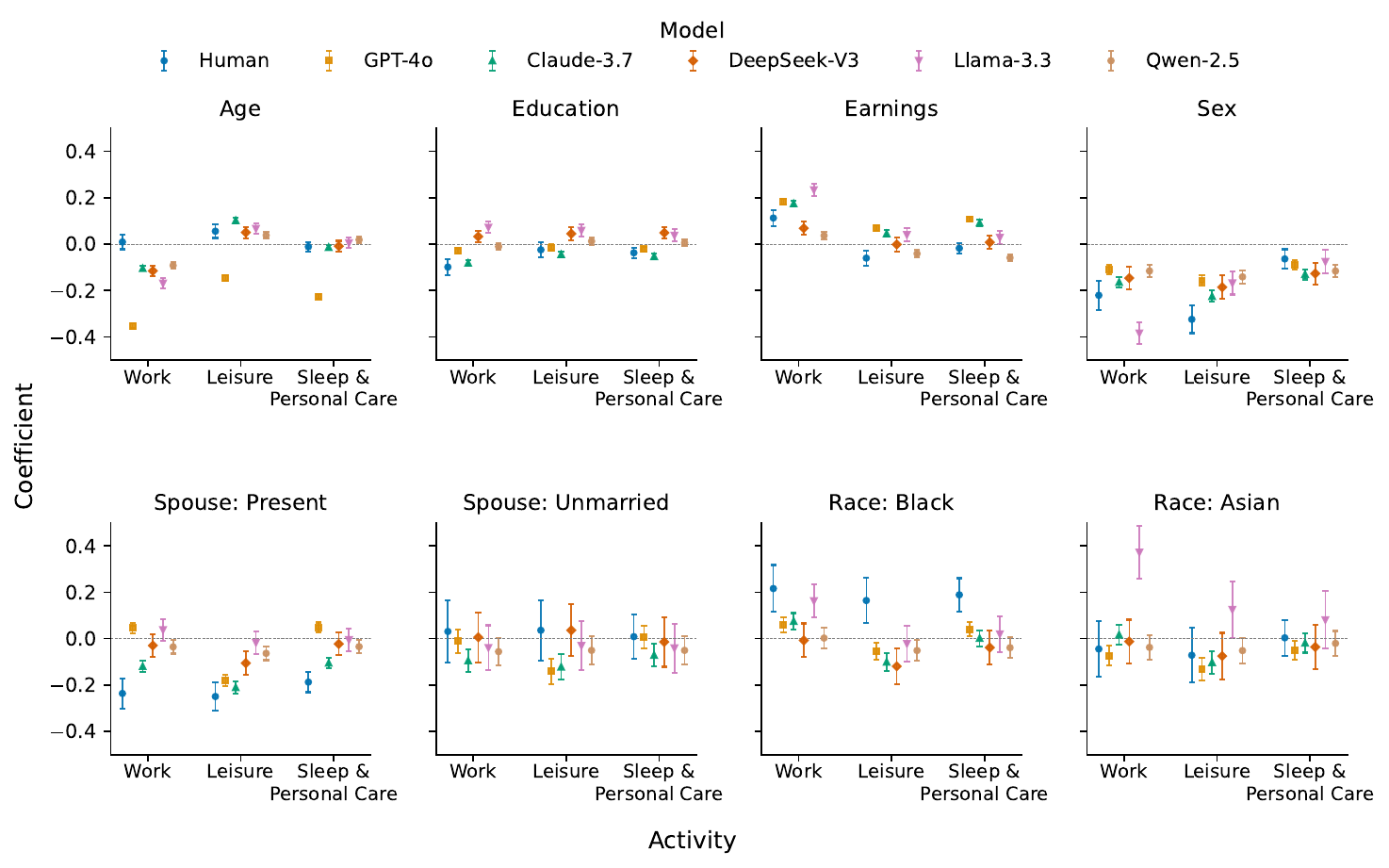}
    \caption{Attribute-Level Comparison of Estimated Decision Weights Between Humans and LLMs}
  \label{fig:results_by_feature}

\end{figure*}

\input{Tables/M_l}

Complementing cosine similarity, Table~\ref{tab:model_misalignment} reports the model-level divergence score $M_l$, which captures the \emph{magnitude} of coefficient gaps from the human baseline. The ranking is broadly consistent with the cosine-based results: Claude-3.7 has the smallest deviation ($M_l=0.094$), while DeepSeek-V3 ($0.126$) and GPT-4o ($0.128$) show moderate gaps. In contrast, Llama-3.3 ($0.180$) and Qwen-2.5 ($0.189$) deviate the most and together account for over half of the total divergence, indicating systematically larger departures from human trade-off weights.

\input{Tables/A_f}

\input{Tables/misalign_black_spouse_v2}

\paragraph{Attribute-Level Misalignment Analysis.}
To further decompose and identify the sources of misalignment between human and LLM decision makers, Figure~\ref{fig:results_by_feature} compares the estimated coefficients feature by feature (with complete estimates in Table~\ref{tab:coef_appendix} in Appendix~\ref{sec: additional_res}). We focus on (i) whether models recover the same \emph{direction} and statistical salience of each weight and (ii) the \emph{magnitude} of divergence from the human baseline. For readability, we omit the estimation results on \textit{Race: Pacific Islander} and \textit{Race: Native American} as sub-figures because the underlying sample sizes are too small to support stable inference, even though these features rank highly in the deviation summary as shown in Table~\ref{tab:feature_misalignment}\footnote{The results with \textit{Race: Pacific Islander} and \textit{Race: Native American} are shown in Figure~\ref{fig:results_by_model} in Appendix~\ref{sec: additional_res}}. Across most covariates, the results do not reveal a single consistent alignment/misalignment pattern: models may track the human sign and significance for some attributes in one activity while diverging in another, consistent with the activity-dependent heterogeneity observed in Figure~\ref{fig:model_heatmap}.

Despite this overall heterogeneity, Figure~\ref{fig:results_by_feature} and Table~\ref{tab:feature_misalignment} highlight two attributes that exhibit particularly large and consistent deviations from the human baseline across models: \textit{Race: Black} and \textit{Spouse: Present}. Table~\ref{tab:feature_misalignment} reports the largest deviation cases within each activity and motivates our focus on these two groups in the main text, while the full coefficient estimates are provided in Table~\ref{tab:coef_appendix} Appendix~\ref{sec: additional_res}. For \textit{Race: Black}, the human coefficients are positive and statistically significant across all three activities (Leisure: $0.164$, Sleep and Personal Care: $0.189$, Work: $0.216$), indicating that relative to the reference category \textit{Other}, Black respondents allocate systematically more time to each focal activity. In contrast, most LLMs substantially attenuate these effects and often reverse the sign, with the largest discrepancies occurring in Leisure (DeepSeek-V3: $-0.119$, Claude-3.7: $-0.100$, GPT-4o: $-0.053$), suggesting that the models encode qualitatively different trade-offs for this group.

A similar pattern holds for \textit{Spouse: Present}. In the human baseline, this coefficient is negative and significant across Leisure ($-0.249$), Sleep and Personal Care ($-0.187$), and Work ($-0.236$), implying that married individuals shift time away from these activities toward the reference category. Several models shrink these negative effects sharply toward zero, and some nearly eliminate them. For example, GPT-4o is close to zero for \textit{Work} and \textit{Sleep and Personal Care}, producing the largest deviations highlighted in Table~\ref{tab:worst_alignment}. 

Under the structural interpretation of coefficients as attribute-dependent reweighting of constrained trade-offs, these discrepancies indicate that LLMs encode systematically different priorities for race and marital status when allocating a fixed time budget. These mechanism-level gaps are not apparent from outcome agreement alone, but become salient when alignment is evaluated through estimated decision weights.

\subsection{Validating XChoice against Reduced-Form Regression}

\begin{figure}[h!]
  \centering
  \includegraphics[width= \linewidth]{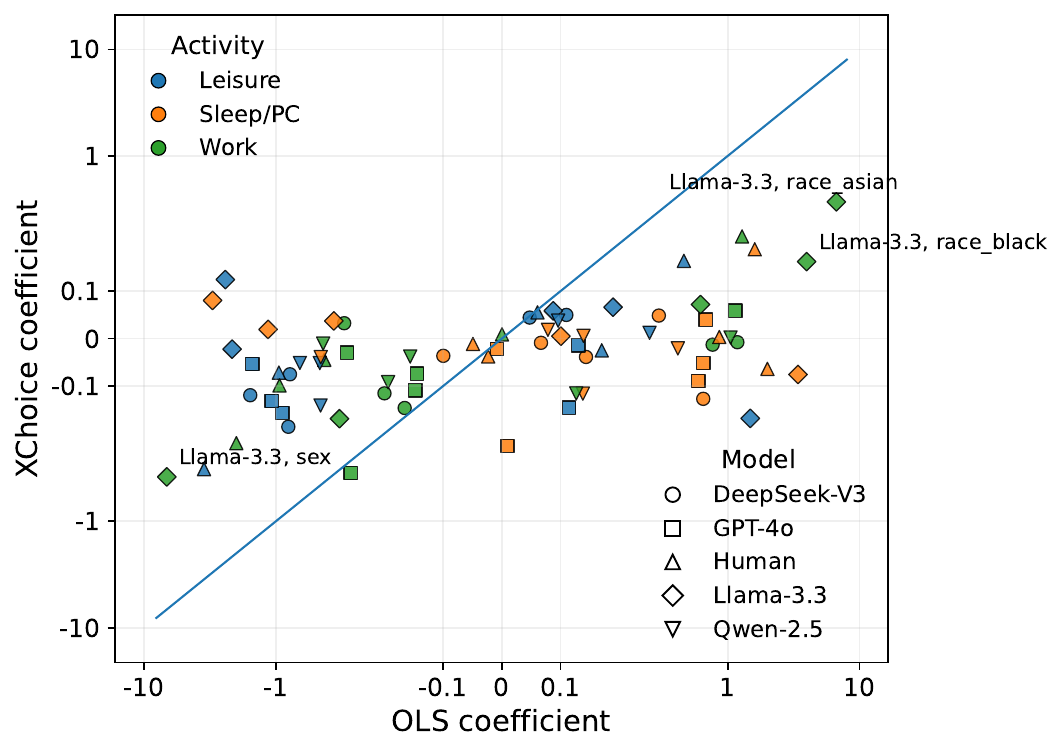}
    \caption{Coefficient Scatter Plot Comparing Reduced-Form Regression and \textsc{XChoice} Estimates}
  \label{fig:ols_vs_xchoice}
  \vspace{-7pt}
\end{figure}

\paragraph{Comparison to Reduced-Form Regressions.}
Figure~\ref{fig:ols_vs_xchoice} compares reduced-form linear regression (OLS) coefficients (x-axis) with \textsc{XChoice} mechanism-based coefficients (y-axis) for the same model–activity–feature tuples, with detailed OLS estimation results reported in Table~\ref{tab:ols_coef_appendix} in Appendix~\ref{sec: additional_res}. If the two approaches yielded similar estimates, points would concentrate around the $45^\circ$ line; instead, we observe substantial dispersion and a clear “shrinkage” pattern: many coefficients that appear extreme under OLS map to much smaller \textsc{XChoice} estimates, with points pulled toward zero on the y-axis. This contrast is especially visible for several demographic features in the \textit{Work} category, where OLS produces large-magnitude effects (far right/left), while \textsc{XChoice} yields more moderate values that reflect trade-offs under the daily time-budget constraint. The most prominent outliers, such as Llama-3.3 on \textit{Race: Asian} and \textit{Race: Black}, highlight cases where OLS suggests strong associations but the structural mechanism implies weaker or qualitatively different reweighting once substitution across activities is accounted for. Overall, the figure suggests that reduced-form regressions can amplify activity-specific correlations, whereas \textsc{XChoice} produces more conservative, mechanism-consistent weights that are less driven by marginal correlations and more aligned with constrained allocation structure.

\input{Tables/invariance}

\paragraph{Robustness/Invariance Analysis.}
To further assess the robustness of \textsc{XChoice} under mild data and policy drift, we conduct an invariance analysis by introducing small shifts in the distributions of earnings, age, race, and spouse status (data shift details in Appendix~\ref{sec:invariance}). Table~\ref{tab:drift_summary} reports the invariance analysis results with three different metrics. Each row reports the drift between $\theta^{(0)}$ (pre-shift estimates) and $\theta^{(1)}$ (post-shift estimates) under one shift; smaller values indicate stronger invariance. Across all four shifts, the structural estimator exhibits consistently lower MAD/RelL2 and near-zero $1-\mathrm{CosSim}$ relative to OLS, implying that the \emph{direction} of the estimated preference vector (i.e., relative trade-offs across covariates) is largely preserved under mild environmental perturbations. This supports the interpretation that \textsc{XChoice} estimation better captures stable latent ``value weights,'' whereas OLS coefficients are more entangled with sample-specific correlational structure and thus drift more under covariate reweighting.

\subsection{Mitigating AI-Human Misalignment with RAG}
Guided by the diagnostics, we focus mitigation on the two most consistently divergent attributes, \textit{Race: Black} and \textit{Spouse: Present}, and target GPT-4o and DeepSeek-V3 as representative closed- and open-source models. As shown in Table~\ref{tab:worst_alignment}, they are the only models with worst-case misalignment in two activities across the two focal subgroups. We then apply a RAG intervention that conditions LLM decisions on evidence retrieved—via embedding similarity—from an academic-article knowledge base built from domain-specific papers, containing 43 and 50 instances on time allocation by marital status and race, respectively, and re-estimate parameters to measure changes in mechanism-level alignment. Implementation details are provided in Appendix~\ref{sec:rag}.

\input{Tables/RAG}

Table~\ref{tab:rag_cossim_results} summarizes how the RAG intervention changes mechanism-level alignment for the two targeted subgroups. For GPT-4o, RAG improves alignment for both \textit{Race: Black} (from $-0.827$ to $-0.782$) and \textit{Spouse: Present} (from $0.276$ to $0.741$), with a particularly large gain for marital status, suggesting that conditioning on retrieved evidence can substantially shift the model’s implied trade-off weights toward the human baseline. For DeepSeek-V3, the effects are mixing: RAG improves alignment on \textit{Race: Black}, where the baseline divergence is relatively large (from $-0.774$ to $-0.673$), but it reduces alignment on \textit{Spouse: Present} (from $0.704$ to $-0.015$), where the baseline alignment is already strong. This contrast suggests that retrieval can be most effective when it targets a genuine gap in the model’s implicit knowledge or priors, while it may be unnecessary, or even disruptive, when the model already encodes human-consistent trade-offs for a given attribute.

\section{Conclusions and Discussions}
In this study, we propose \textsc{XChoice}, an explainable framework for evaluating AI-human alignment in constrained decision making using mechanism-based, structural metrics. Rather than relying on outcome-level agreement alone, \textsc{XChoice} recovers and compares interpretable decision parameters that capture how humans and LLMs weight attributes, satisfy constraints, and resolve trade-offs. Through a case study on Americans’ daily time allocation with ATUS as human ground truth, our results show that model alignment is highly heterogeneous across activities and features, and that several salient discrepancies are concentrated in specific attributes and subgroups such as \textit{Race: Black} and \textit{Spouse: Present}, which would be difficult to diagnose using accuracy-style metrics.

Beyond diagnosis, \textsc{XChoice} also supports informed misalignment mitigation. We show how discrepancy signals guide targeted interventions by applying a RAG strategy to GPT-4o and DeepSeek-V3 and re-estimating parameters after conditioning on retrieved evidence. Results illustrate both promise and limits: retrieval can improve alignment where models exhibit clear mechanism gaps, but may offer little benefit or even degrade alignment when baseline estimates are already close to human behavior. We also benchmark robustness via an invariance analysis under environment shifts and contrast it with reduced-form baselines, consistent with the Lucas-critique motivation. Overall, our findings suggest alignment hinges on whether models learn human-like trade-off weights that encode stable priorities, not just whether they match observed choices in a single context.

\newpage 

\section*{Limitations}

Our work has several limitations. First, \textsc{XChoice} is designed for \emph{constrained decision making} problems where choices can be formalized with an explicit objective, a feasible set, and interpretable trade-offs. While this class of decisions is common and consequential, LLMs are deployed in a much broader range of settings (e.g., open-ended generation, multi-step reasoning, negotiation, and interactive assistance) that may not admit a clean constrained-optimization representation, and extending mechanism-based alignment metrics to those contexts remains an open direction. Second, our approach inherits limitations from structural estimation. The recovered parameters depend on the assumed model form and normalization choices, and they summarize behavior within the scope of observed covariates and constraints; misspecification, omitted variables, or heterogeneity beyond what the model captures can bias estimates and complicate interpretation. Moreover, invariance is best viewed as an empirical property rather than a guarantee: structural parameters may still drift under large regime shifts or when preferences and institutions evolve. These limitations motivate future work on more flexible structural formulations, richer sources of behavioral variation, and broader validation across decision environments. 

Although \textsc{XChoice} is intended as an evaluation tool, mechanism-based diagnostics could be misused to engineer models that better imitate human decision patterns for manipulative or discriminatory purposes, especially when applied to sensitive attributes or subgroups. In addition, subgroup-level analyses may be over-interpreted as normative claims about how groups \emph{should} behave, or may reinforce stereotypes if reported without appropriate context and uncertainty. To mitigate these risks, we emphasize that our metrics are descriptive rather than prescriptive, report uncertainty, and recommend avoiding deployment decisions based solely on subgroup-specific coefficients without complementary qualitative and domain review.



\bibliography{custom}

\clearpage
\appendix

\section*{Appendix}
\label{sec:appendix}

\section{Prompts to LLMs}
\label{sec: prompt}
\begin{tcolorbox}[
    breakable,
    colback=gray!12,
    colframe=black!60,
    boxrule=0.4pt,
    arc=2pt,
    left=2mm,
    right=2mm,
    top=1mm,
    bottom=1mm,
    fontupper=\small\ttfamily
]
\textbf{System Instruction:} 
You are an American making daily time allocation decisions across three activities: 
work, leisure, and other. Your goal is to maximize overall happiness within the 
1,440 minutes available each day.

\medskip

\textbf{User Prompt Template:} 
You are a \{Gender Identity\}, \{Age\} years old, ethnically identified as \{Race\}. Your highest level of education is 
\{Educational Level\}, and your weekly income is \{Income\}. Based on this background, 
how would you allocate your time across Work, Leisure, Sleep and Personal Care, and Other in a typical day? 
Please answer in the format: [Work, Leisure, Sleep and Personal Care, Other], using numbers to indicate minutes.
\end{tcolorbox}

\captionsetup{type=listing}
\captionof{lstlisting}{Prompt template for personalized LLM-based agents.}

\section{Data Preprocessing}
\label{sec: data_preprocessing}

\input{Tables/summary_stat}

This section describes the data preprocessing procedures used to transform the raw 
survey data into a structured format suitable for subsequent analysis. The original 
ATUS extract contains 8,548 respondents. We first restrict the sample to individuals 
with non-missing information on key covariates, dropping observations coded as 
``not in universe'' or with missing earnings (e.g., \texttt{EARNWEEK} = 99999.99). 
After these restrictions, the final analytic sample consists of 4,307 respondents.

We then construct a set of harmonized covariates. Race is recoded from the original 
multi-category \texttt{RACE} variable into four single-race indicators (Black, Native 
American, Asian, and Pacific Islander), using White-only respondents as the reference 
group. Educational attainment is collapsed from the detailed \texttt{EDUCYRS} codes 
into an ordered four-level variable (\texttt{edu}) distinguishing high school or less, 
some college, bachelor's degree, and any postgraduate education. Sex is recoded into a 
binary indicator for male (with female as the reference category), and household 
partnership status is represented by two indicators for spouse present and unmarried 
partner present (with no spouse or partner as the reference category). 

In addition to these attributes, we retain respondents' time allocation across four 
activity categories—work, leisure, sleep and personal care, and other—each measured in 
minutes per day. The resulting dataset provides a clean and interpretable set of 
variables for the subsequent empirical analysis and LLM-based simulations.

\paragraph{Sex.}
Respondents' sex is recorded in the variable \texttt{SEX} (\texttt{01} = male, 
\texttt{02} = female, \texttt{99} = NIU). We exclude observations coded as NIU 
(\texttt{SEX} = 99) and recode this variable into a binary indicator. Specifically, 
we define a variable \emph{Male} that takes the value 1 if the respondent is coded as 
male (\texttt{SEX} = 01) and 0 if coded as female (\texttt{SEX} = 02), treating 
female as the reference category. This binary indicator is used as the sex covariate 
in the subsequent analysis.

\paragraph{Race.}
The original dataset reports respondents' race using the variable \texttt{RACE}, a 
multi-category code that distinguishes single-race and multiracial combinations (e.g., 
\texttt{0100} = White only, \texttt{0110} = Black only, \texttt{0120} = American 
Indian/Alaskan Native, \texttt{0131} = Asian only, \texttt{0132} = Hawaiian/Pacific 
Islander only, and codes $\geq \texttt{0200}$ for various multiracial combinations). 
To focus on clearly defined single-race groups and avoid small, heterogeneous multiracial 
cells, we restricted the sample to respondents with a single reported race by dropping all 
observations with codes greater than or equal to \texttt{200} (\texttt{drop if race >= 200} 
in Stata). We then used ``White only'' as the baseline category and constructed four binary 
indicator variables---\emph{Race: Black}, \emph{Race: Native American}, \emph{Race: Asian}, 
and \emph{Race: Pacific Islander}---which take the value 1 if the respondent belongs to the 
corresponding single-race group and 0 otherwise. These indicators enter the subsequent 
models as race covariates, with White as the omitted reference group.

\paragraph{Income.}
Weekly earnings are measured using the variable \texttt{EARNWEEK}, which reports 
respondents' usual weekly income. In the original data, missing values are coded as 
\texttt{99999.99}. We treat this code as missing and drop all such observations from 
the analysis (\texttt{EARNWEEK = 99999.99}). The resulting non-missing values of 
\texttt{EARNWEEK} are used as the weekly income covariate in our models.

\paragraph{Education.}
The survey records respondents' educational attainment using the variable \texttt{EDUCYRS}, 
which distinguishes detailed schooling levels (e.g., grades 1--12, one to four years of 
college, bachelor's degree, and various graduate degrees). For ease of interpretation and 
to avoid sparsely populated categories, we collapsed \texttt{EDUCYRS} into an ordered 
four-level variable \texttt{edu}. Specifically, we coded \texttt{edu} = 1 for respondents 
with at most a high school education ($\texttt{EDUCYRS} \leq 112$, grades 1--12), 
\texttt{edu} = 2 for those with some college but no bachelor's degree 
($112 < \texttt{EDUCYRS} \leq 216$), \texttt{edu} = 3 for respondents with a bachelor's 
degree ($\texttt{EDUCYRS} = 217$), and \texttt{edu} = 4 for respondents with any 
postgraduate education ($\texttt{EDUCYRS} > 300$), including master's, professional, 
and doctoral degrees. This four-level variable is used as the education covariate in 
subsequent analyses.

\paragraph{Spouse/partner status.}
We measure household partnership status using the variable \texttt{SPOUSEPRES}, which 
indicates whether a spouse or unmarried partner is present in the household 
(\texttt{01} = spouse present, \texttt{02} = unmarried partner present, 
\texttt{03} = no spouse or unmarried partner present, \texttt{99} = NIU). 
We exclude observations coded as NIU (\texttt{SPOUSEPRES} = 99) and use 
``no spouse or unmarried partner present'' (\texttt{SPOUSEPRES} = 03) as the 
reference category. We then construct two binary indicator variables, 
\emph{Spouse present} and \emph{Unmarried partner present}, which take the value 
1 when \texttt{SPOUSEPRES} equals 01 or 02, respectively, and 0 otherwise. 
These indicators enter the models as covariates capturing household partnership status.

\paragraph{Feature scaling and coding.}
Before estimation, we standardize and encode all covariates in a consistent way. 
On the continuous side, we treat respondents' age, educational attainment, and 
weekly earnings as continuous predictors and include \texttt{age}, the four-level 
education variable \texttt{edu}, and weekly earnings \texttt{EARNWEEK}. These 
three variables are standardized to have mean zero and unit variance using the 
sample mean and standard deviation from the final analytic sample. This scaling 
improves numerical stability and makes the corresponding coefficients more 
comparable in magnitude across covariates.

Binary and categorical variables are kept on an interpretable 0/1 scale and are 
not standardized. Sex enters the models as a binary indicator \emph{Male}, coded 
as 1 for respondents coded as male in \texttt{SEX} and 0 for those coded as 
female (the reference category). Household partnership status, originally recorded 
in \texttt{SPOUSEPRES}, is represented by two indicator variables, \emph{Spouse 
present} and \emph{Unmarried partner present}, with “no spouse or unmarried partner 
present” serving as the baseline category. Race, originally coded in \texttt{RACE}, 
is represented by four single-race indicators—\emph{Race: Black}, \emph{Race: 
Native American}, \emph{Race: Asian}, and \emph{Race: Pacific Islander}—with 
“White only” as the omitted reference group. All standardized continuous variables 
and dummy-coded indicators are concatenated into a single design matrix, and we 
include an explicit intercept term for use in the subsequent empirical analysis 
and LLM-based simulations.

\section{Descriptive Statistics}
\label{sec:data_summary}

\input{Tables/est_res}

Table~\ref{tab:summary_stat} reports descriptive statistics for respondent attributes and 
time-use outcomes in the final analytic sample of 4{,}307 individuals. By construction, 
the continuous attributes (age, education, and weekly earnings) are standardized to 
have mean zero and unit variance. The binary attributes indicate that roughly half of 
the sample is male (mean $\approx 0.51$), just over half live with a spouse 
($\approx 0.54$), and a small fraction live with an unmarried partner ($\approx 0.06$). 
The racial composition is predominantly White, with comparatively small shares of 
Black, Asian, Native American, and Pacific Islander respondents. On average, 
respondents allocate about 247 minutes per day to work, 266 minutes to leisure, 
578 minutes to sleep and personal care, and 349 minutes to other activities, with 
substantial cross-individual variation in both work and non-work time.

\section{Additional Details of Experimental Setup}
\label{sec:exp_models}
\paragraph{LLMs under evaluation.}
We evaluate a diverse set of LLMs to demonstrate that \textsc{XChoice} applies across major model families and capability levels. 
For closed-source systems, we include \texttt{gpt-4o-2024-08-06} (\emph{GPT-4o}) and \texttt{claude-3-7-sonnet-20250219} (\emph{Claude-3.7}). We also assess leading open source models, including \texttt{Qwen/Qwen2.5-72B-Instruct} (\emph{Qwen-2.5}), \texttt{meta-llama/Llama-3.3-70B-Instruct} (\emph{Llama-3.3}), and \texttt{deepseek-ai/DeepSeek-V3} (\emph{DeepSeek-V3}). 
When using provider-specific endpoints, we report the underlying base model name for clarity and reproducibility.

\section{Additional Experimental Results}
\label{sec: additional_res}

\begin{figure*}[h!]
  \centering
  \includegraphics[width= \textwidth]{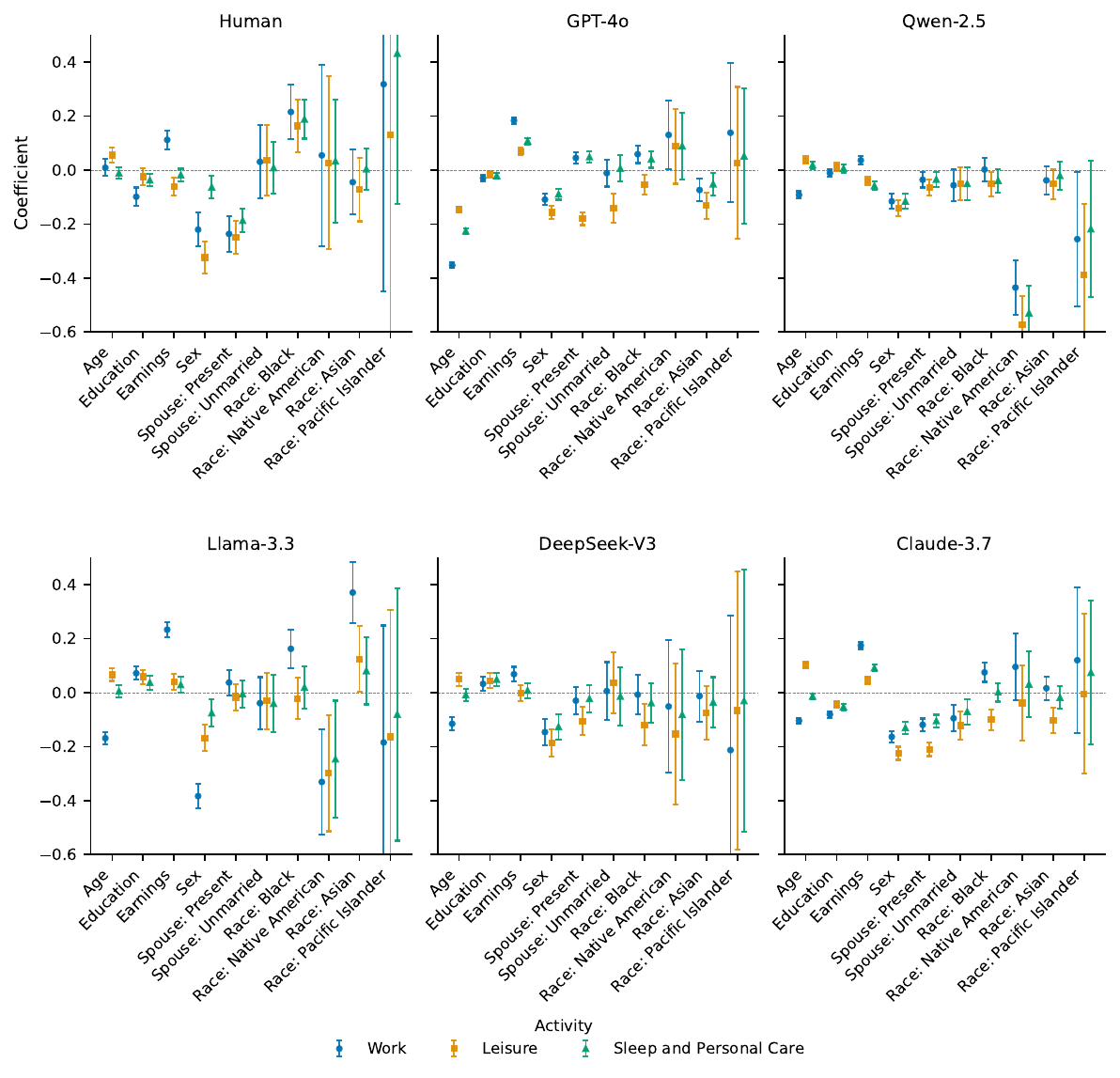}
    \caption{Estimated Decision-Weight Profiles by Model (Across Activities and Attributes)}
  \label{fig:results_by_model}
\end{figure*}

\begin{figure*}[h!]
  \centering
  \includegraphics[width= \textwidth]{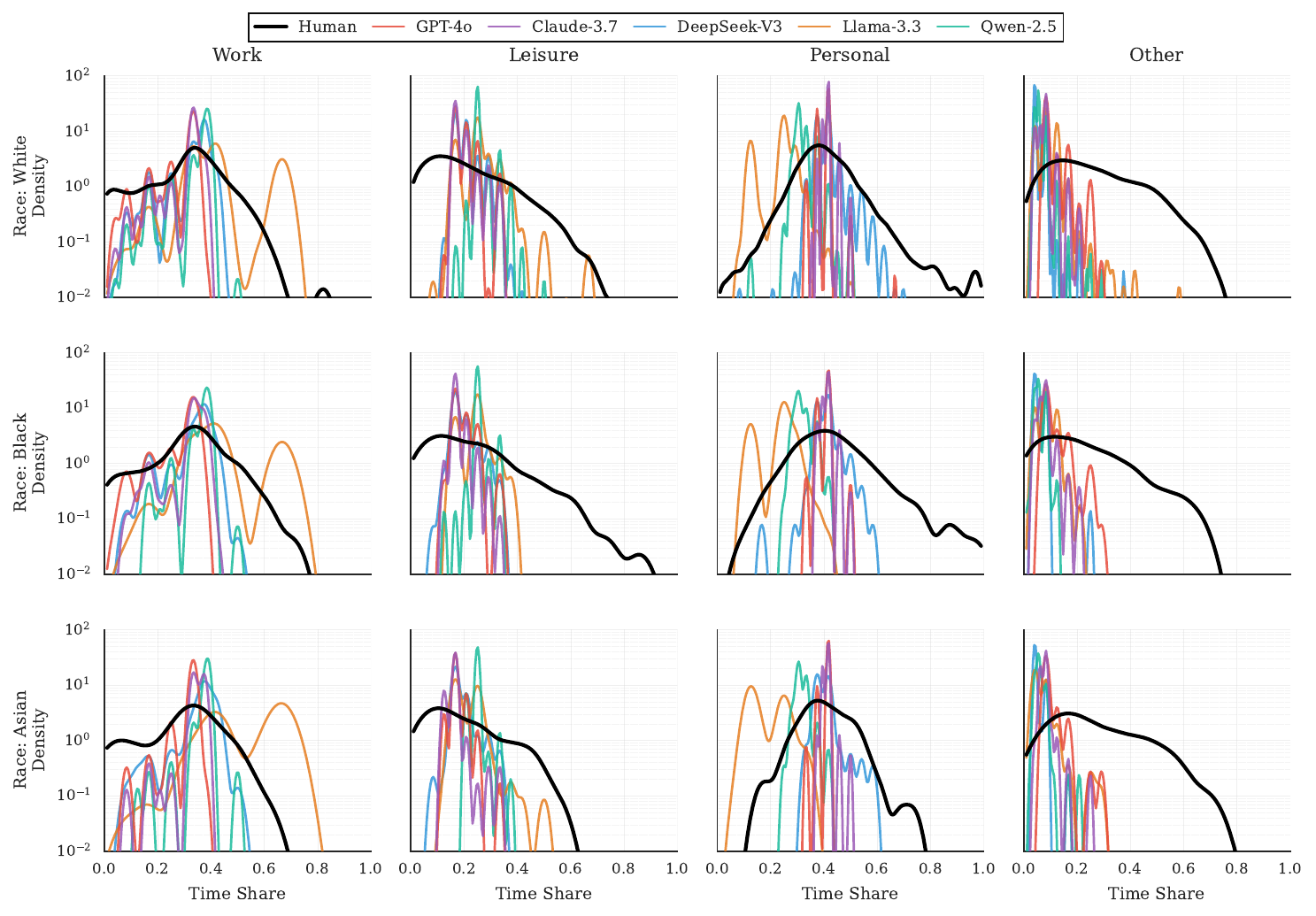}
    \caption{Time-share distributions across activities by race group. 
    Each column corresponds to one activity (Work, Leisure, Personal, Other), and each row 
    to a race group. Within each panel, we compare the empirical distribution of human 
    time shares (black) with time-share patterns produced by different LLM-based models, 
    using a log-scaled density axis.}

  \label{fig:time_share_race}
\end{figure*}

\begin{figure*}[h!]
  \centering
  \includegraphics[width= \textwidth]{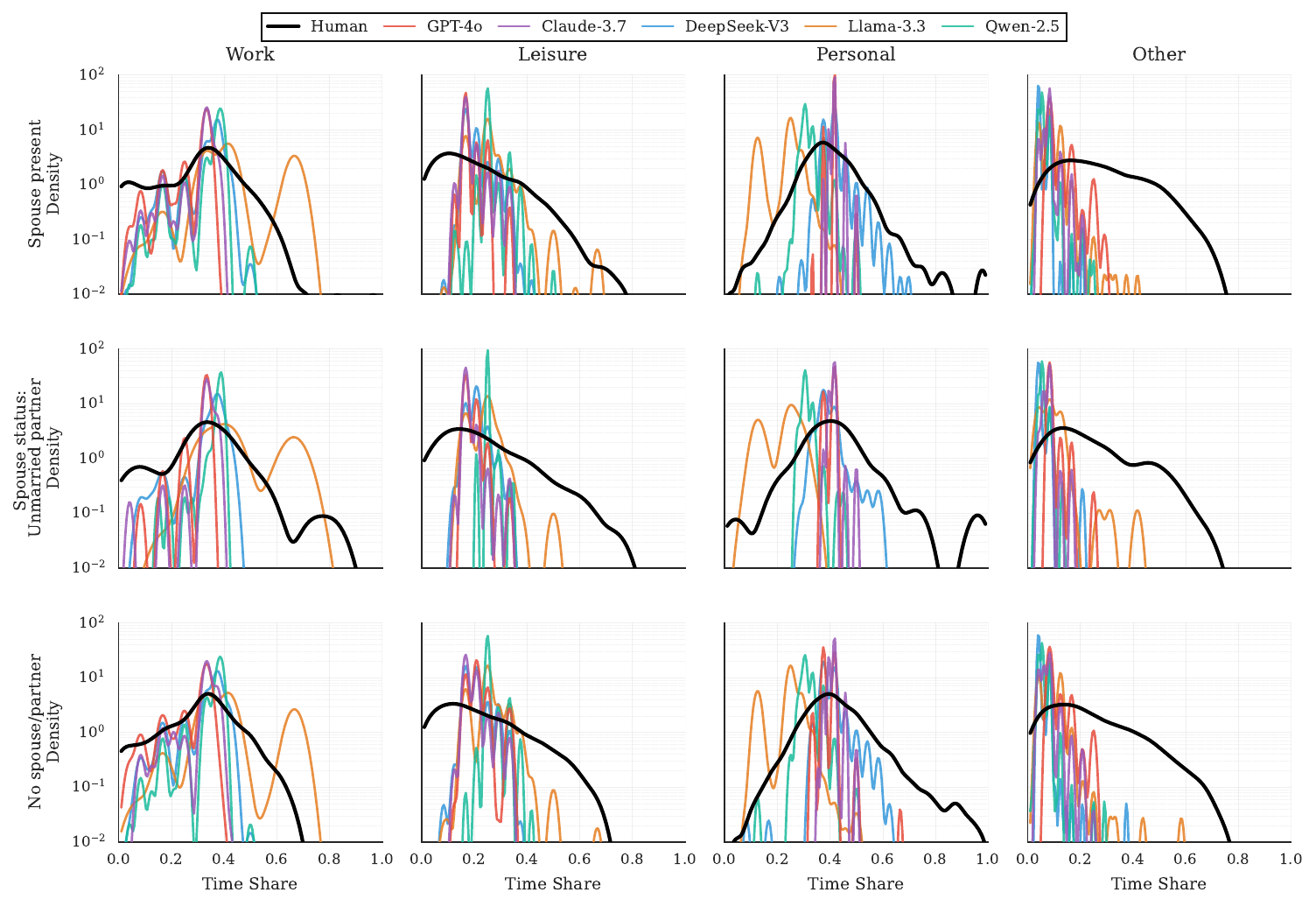}
    \caption{Time-share distributions across activities by household partnership status. 
    Each column corresponds to one activity (Work, Leisure, Personal, Other), and each row 
    to a spouse/partner status category (spouse present, unmarried partner, no spouse/partner). 
    Within each panel, we compare the empirical distribution of human time shares (black) 
    with time-share patterns produced by different LLM-based models, using a log-scaled 
    density axis.}

  \label{fig:time_share_spouse}
\end{figure*}

\input{Tables/ols_res}

\paragraph{Full coefficient estimates.}
Table~\ref{tab:coef_appendix} reports the full coefficient estimates for each activity-specific model, including the human baseline and all evaluated LLMs. Coefficients are presented as point estimates with 95\% confidence intervals, and statistically significant estimates ($p<0.05$) are boldfaced. We include this table as supplementary evidence for the attribute-level comparisons in the main text (Figure~\ref{fig:results_by_feature} and Table~\ref{tab:feature_misalignment}), enabling readers to inspect the exact magnitudes, signs, and uncertainty of each demographic factor across activities and models.

\paragraph{Decision-weight profiles by model.}
Figure~\ref{fig:results_by_model} complements Figure~\ref{fig:results_by_feature} by reorganizing the same coefficient estimates at the \emph{model} level. Each panel corresponds to one decision maker (Human or an LLM), and the x-axis lists features, while colors indicate activities (Work, Leisure, and Sleep/Personal Care) with 95\% confidence intervals. Unlike the main-text attribute-level plot, we include \textit{Race: Pacific Islander} and \textit{Race: Native American} here for completeness; the wide intervals and unstable estimates illustrate why these categories are omitted from the main results due to limited sample size and lack of meaningful statistical precision. Overall, the model-level view makes it easier to inspect how each model reweights attributes across activities and where uncertainty or sign changes concentrate.

\paragraph{Distributional view of misalignment identification.}
Figures~\ref{fig:time_share_race} and Figure~\ref{fig:time_share_spouse} provide a distributional view of the misalignment identified by the coefficient-based diagnostics. Across both race and partnership-status strata, human time shares (black curves) are noticeably more diffuse, with heavier tails that reflect heterogeneous schedules and constraints. In contrast, LLM-generated shares are highly concentrated, showing clear heaping at a few focal values, which is consistent with stereotyped allocations such as round-hour schedules. A prominent pattern is that models systematically under-allocate the residual category \textit{Other}: compared with the human distributions, LLM densities collapse toward near-zero time shares and rarely reproduce the wider support and long right tail observed in the survey data. This suggests that models tend to over-commit time to the salient categories (Work/Leisure/Personal) while failing to capture the diversity of non-core activities that occupy substantial portions of the day for many individuals.

The subgroup panels also highlight why \textit{Race: Black} and \textit{Spouse: Present} emerge as salient divergence sources in the mechanism estimates. In the human data, the black curves shift in shape across strata, indicating that demographic attributes are associated with meaningful changes in how time is distributed across activities (including tail behavior and multimodality). By comparison, many LLM curves vary only weakly across rows, implying limited sensitivity to subgroup-specific trade-offs even when demographic information is provided. Put differently, the models often produce similar “template” schedules across groups, compressing within-group variation and attenuating cross-group differences. This distributional evidence is consistent with the coefficient misalignment results: the models do not merely miss outcomes at the margin, but encode different prioritization patterns for these subgroups, especially in how they allocate residual time and how strongly demographic attributes reshape the allocation trade-offs.

\paragraph{OLS estimation results.}
Table~\ref{tab:ols_coef_appendix} reports reduced-form OLS estimates for each activity and decision maker, including the human baseline and LLM-generated decisions. For each activity, we regress the observed or model-produced time share on the same set of demographic covariates (standardized age, education, weekly earnings, plus indicators for sex, spouse or partner status, and race), and report coefficients with 95\% confidence-interval margins. These coefficients summarize conditional associations and do not enforce the time-budget adding-up constraint across activities, so they should not be interpreted as trade-off weights. Instead, we use them as a conventional baseline to contrast against \textsc{XChoice}’s mechanism-based estimates and to support robustness checks under mild environment shifts. The table also highlights that coefficients for sparsely represented groups, especially Native American and Pacific Islander, can be highly variable and imprecise, which aligns with our decision to omit these categories from the main-text attribute-level visualization.

\section{Invariance Analysis Metrics and Experimental Design}
\label{sec:invariance}

We test parameter invariance by comparing baseline estimates $\theta^{(0)}$ to estimates $\theta^{(1)}$ obtained after applying mild counterfactual data shifts. Drift is quantified by three lower-is-better metrics:
\begin{align}
\mathrm{MAD} 
&= \frac{1}{K}\sum_{i=1}^{K}\left|\theta^{(1)}_{i}-\theta^{(0)}_{i}\right|, \\
\mathrm{RelL2}
&= \frac{\left\|\theta^{(1)}-\theta^{(0)}\right\|_{2}}{\left\|\theta^{(0)}\right\|_{2}}, \\
1-\mathrm{CosSim}
&= 1-\frac{\theta^{(1)}\cdot\theta^{(0)}}{\left\|\theta^{(1)}\right\|_{2}\left\|\theta^{(0)}\right\|_{2}}.
\end{align}

MAD measures average absolute coefficient change, RelL2 measures relative change in $\ell_2$ magnitude, and $1-\mathrm{CosSim}$ measures directional rotation (near $0$ implies preserved relative trade-offs).

\paragraph{Data shift design.}
We implement four shifts that emulate small policy-like perturbations and compositional changes:
\begin{itemize}[leftmargin=*]
    \item \textbf{Earnings quantile lift (magnitude 0.5).} Let $q_{0.5}$ be the median of Weekly earnings (\texttt{earnweek}). We only lift the bottom half by adding $0.5\cdot \mathrm{sd}(\texttt{earnweek})$ to all individuals with $\texttt{earnweek}\le q_{0.5}$, leaving the upper half unchanged. This changes the \emph{shape} of the earnings distribution (not a pure location shift).
    \item \textbf{Age band shift (magnitude $0.1$).} We randomly select a fraction $p=0.1$ of individuals aged $25$--$34$ and increase their continuous \texttt{age} by $10$ years, effectively moving them into the $35$--$44$ range. This produces a localized age redistribution and can modify correlations between age and other covariates (e.g., spouse/education/earnings).
    \item \textbf{Race mix shift (Asian $+2$pp, White $-2$pp).} After discretizing age into \texttt{age\_band}, we perform a mild raking-style resampling within each (\texttt{age\_band}, \texttt{sex}) stratum to increase race code $131$ (Asian) by $0.02$ in share and decrease race code $100$ (White; baseline) by $0.02$, keeping other categories unchanged up to normalization. This is a small compositional (mixture) shift that preserves the stratification structure.
    \item \textbf{Spouse mix shift (present $+3$pp in ages $30$--$44$).} Using \texttt{age\_band} bins $(18,25,30,35,45,55,65,100)$, we apply raking-style resampling \emph{only} within age bands covering $30$--$44$ to increase \texttt{spousepres}$=1$ (present) by $0.03$ and decrease \texttt{spousepres}$=3$ (baseline) by $0.03$. Other age bands are left unchanged. This creates a targeted, policy-like compositional perturbation rather than a global reweighting.
\end{itemize}



\section{RAG Design for Misalignment Mitigation}
\label{sec:rag}

To analyze alignment at the attribute (and subgroup) level, we compare how each model weights the same feature across activities.
For a given feature $f$ (e.g., \textit{Race: Black} or \textit{Spouse: Present}), we form the activity-specific coefficient vector
$\hat{\theta}_{m,f} = (\hat{\theta}^{L}_{m,f}, \hat{\theta}^{W}_{m,f}, \hat{\theta}^{S}_{m,f})$ for each model $m$, and likewise
$\hat{\theta}_{H,f}$ for the human baseline.
We then quantify alignment using cosine similarity:
\[
CosSim(\hat{\theta}_{H,f}, \hat{\theta}_{m,f})
= \frac{\hat{\theta}_{H,f}^\top \hat{\theta}_{m,f}}
{\|\hat{\theta}_{H,f}\|\,\|\hat{\theta}_{m,f}\|}.
\]
This metric captures whether humans and LLMs exhibit similar \emph{directional} trade-off patterns for feature $f$ across activities, independent of overall scale, and enables systematic diagnosis of which attributes drive mechanism-level divergence.

We employ a Retrieval-Augmented Generation (RAG) pipeline to enhance LLM predictions of time allocation by grounding model outputs in empirical research findings. The following subsections describe the construction of our knowledge base, the retrieval mechanism, and the prompt design.

\subsection{Knowledge Base Construction}
\label{subsec:knowledge_base}

We construct domain-specific knowledge bases from peer-reviewed research on time allocation behavior. Each knowledge base contains structured instances extracted from empirical studies, capturing findings about how demographic factors influence daily time use patterns.

\paragraph{Marriage Knowledge Base.} This knowledge base comprises 50 instances derived from research papers examining the relationship between marital status and time allocation. Topics include household decision-making dynamics, work hours by marital status, couple time and family activities, the impact of children on time allocation, leisure time patterns, personal care behaviors, and caregiving responsibilities.

\paragraph{Race Knowledge Base.} This knowledge base contains 43 instances from studies investigating racial and ethnic disparities in time use. Topics include leisure-time physical activity disparities, work-related activity patterns, time penalties faced by racial minorities, employment barriers, structural factors affecting time allocation, gender differences in household activities, and sleep duration variations across racial groups.

\subsection{Embedding and Retrieval}
\label{subsec:retrieval}

We employ dense retrieval to identify relevant knowledge instances for each respondent persona.

\paragraph{Embedding Generation.} We generate embeddings for both knowledge instances and respondent personas using OpenAI's \texttt{text-embedding-3-large} model, which produces 3,072-dimensional dense vector representations. Each knowledge instance is embedded based on its textual content, while respondent personas are embedded using a structured sentence template:

\begin{quote}
\textit{``A [age]-year-old [gender] [race] with [education], [spouse status], earning \$[income] per week.''}
\end{quote}

\paragraph{Retrieval Mechanism.} For each respondent, we compute cosine similarity between the persona embedding and all knowledge instance embeddings. We retrieve the top-$k$ most similar instances, where $k=3$ in our experiments.

\subsection{Demographic Variable Encoding}
\label{subsec:demographics}

We encode demographic variables from the ATUS dataset into natural language descriptions. Gender is encoded as male or female. Education level is categorized into four groups: no college education, some college without a bachelor's degree, bachelor's degree, and advanced degree. Race categories include White, Black, American Indian or Alaskan Native, Asian, and Native Hawaiian or Pacific Islander. Spouse presence is encoded as having a spouse, having an unmarried partner, or having no spouse or unmarried partner.

\subsection{Model Configuration}
\label{subsec:model_config}

We evaluate two LLMs in our experiments: GPT-4o and DeepSeek-V3. Both models are configured with a temperature of 0.1 to promote deterministic outputs, a maximum token limit of 1,024, and top-$p$ sampling set to 1. These settings balance output consistency with sufficient generation capacity for the structured JSON response.

\subsection{Illustrative Examples}
\label{subsec:examples}

We present examples demonstrating how the RAG pipeline retrieves contextually relevant knowledge for different demographic profiles. For each example, we show the respondent's demographic information, the constructed persona sentence used for embedding, and the full text of the three retrieved knowledge instances.

\subsubsection{Marriage Experiment Examples}

\paragraph{Example 1: Married Individual.}

\noindent\textbf{Demographics:} 53-year-old Black male with an advanced degree, married, weekly income of \$1,173.

\noindent\textbf{Persona sentence:} \textit{``A 53-year-old male Black with an advanced degree, a spouse, earning \$1173.00 per week.''}

\noindent\textbf{Retrieved knowledge instances:}
\begin{enumerate}
    \item \textit{``Expected wage profiles at the time of marriage strongly impact household decision-making weights, favoring the spouse with a higher expected wage.''}
    \item \textit{``Dual-earner households exhibit more frequent adjustments in weights due to wage shocks compared to single-earner households.''}
    \item \textit{``Understanding how marital status and wage profiles affect intra-household decision-making can inform interventions designed to improve family welfare and resource allocation.''}
\end{enumerate}

\paragraph{Example 2: Unmarried Individual.}

\noindent\textbf{Demographics:} 39-year-old White female with some college education but no bachelor's degree, no spouse or unmarried partner, weekly income of \$240.

\noindent\textbf{Persona sentence:} \textit{``A 39-year-old female White with some college education but no bachelor's degree, no spouse or unmarried partner, earning \$240.00 per week.''}

\noindent\textbf{Retrieved knowledge instances:}
\begin{enumerate}
    \item \textit{``Married individuals, particularly husbands, tend to work longer hours compared to their unmarried peers.''}
    \item \textit{``Married individuals often have more stable employment patterns, prioritizing job security to support their families. Unmarried individuals tend to engage in more part-time work or gig economy jobs, reflecting a more flexible approach to employment.''}
    \item \textit{``Leisure hours are relatively similar between genders, with slight differences in how they allocate time to leisure activities.''}
\end{enumerate}

\subsubsection{Race Experiment Examples}

\paragraph{Example 3: Black Individual.}

\noindent\textbf{Demographics:} 53-year-old Black male with an advanced degree, married, weekly income of \$1,173.

\noindent\textbf{Persona sentence:} \textit{``A 53-year-old male Black with an advanced degree, a spouse, earning \$1173.00 per week.''}

\noindent\textbf{Retrieved knowledge instances:}
\begin{enumerate}
    \item \textit{``Leisure activities account for 53.8\% of the time for urban African American adolescents, which is higher than in other postindustrial populations. Significant engagement in leisure activities includes TV viewing (17.5\%) and talking (11.1\%). Boys reported more leisure time than girls among the urban African American adolescents.''}
    \item \textit{``Urban African American adolescents spent 21.6\% of their time on productive activities, with only 4.5\% dedicated to homework, which is less than their suburban peers.''}
    \item \textit{``Urban African American adolescents spent 3.99\% of their time on household chores and 7.89\% on personal maintenance, with girls reporting more time in these activities.''}
\end{enumerate}

\paragraph{Example 4: Non-Black Individual.}

\noindent\textbf{Demographics:} 30-year-old White female with a bachelor's degree, unmarried partner, weekly income of \$1,730.76.

\noindent\textbf{Persona sentence:} \textit{``A 30-year-old female White with a bachelor's degree, an unmarried partner, earning \$1730.76 per week.''}

\noindent\textbf{Retrieved knowledge instances:}
\begin{enumerate}
    \item \textit{``Women, particularly Spanish-speaking Hispanics, report higher engagement in heavy household chores compared to their White counterparts, suggesting a gendered aspect of time allocation in household responsibilities.''}
    \item \textit{``Non-Whites report higher levels of work-related physical activity (WRPA), with Spanish-speaking Hispanics engaging in strenuous job-related activities most frequently. Whites report the lowest levels of WRPA compared to other racial and ethnic groups.''}
    \item \textit{``Urban African American adolescents spent 3.99\% of their time on household chores and 7.89\% on personal maintenance, with girls reporting more time in these activities.''}
\end{enumerate}

\end{document}

%% file: Tables/method_comparison.tex
\begin{table*}[h]
\centering
\caption{Comparison of three alignment evaluation families. \textsc{XChoice} recovers interpretable latent preference weights via an explicit choice mechanism, and supports generalization to decisions under new environments}
\small
\setlength{\tabcolsep}{6pt}
\renewcommand{\arraystretch}{1.15}
\begin{tabular}{p{2.3cm} p{3.2cm} p{3.8cm} p{4.1cm}}
\toprule
\textbf{Aspect} &
\textbf{Outcome Metrics} &
\textbf{Reduced-Form} &
\textbf{\textsc{XChoice}} \\
\midrule
\textbf{Measurement} &
Match rate on decisions (AI vs.\ human) &
Associations between inputs and observed choices &
Latent preference weights (trade-offs) in a utility model \\
\addlinespace[1pt]
\textbf{Explainability} &
Low (Black-box) &
Moderate (Correlational) &
High (Mechanistic / Causal) \\
\addlinespace[1pt]
\textbf{Constraints} &
Not applicable &
Must be imposed explicitly &
Built in via model structure \\
\addlinespace[1pt]
\textbf{Generalizability} &
Not generalize beyond observed outcomes &
Sensitive to distribution shifts &
Transfers to new environments; stable preference weights\\
\bottomrule
\end{tabular}
\label{tab:alignment_methods}
\vspace{-15pt}
\end{table*}

%% file: Tables/M_l.tex
\begin{table}[h!]
\centering
\caption{Model-level alignment deviation between humans and LLM-based models.
For each model $l$, $M_l$ is the mean absolute deviation of its coefficients
from the human baseline (lower is better); ``Share'' reports the fraction of total deviation
$\sum_l M_l$ accounted for by each model.}
\label{tab:model_misalignment}
\small
\begin{tabular}{l r r}
\toprule
Model & Mean $M_l$ & Share (\%) \\
\midrule
Qwen-2.5     & 0.189 & 26.4 \\
Llama-3.3   & 0.180 & 25.1 \\
GPT-4o      & 0.128 & 17.8 \\
DeepSeek-V3 & 0.126 & 17.6 \\
Claude-3.7      & 0.094 & 13.1 \\
\bottomrule
\end{tabular}
\end{table}

%% file: Tables/A_f.tex
\begin{table}[h!]
\centering
\caption{Feature-wise alignment deviation between human and LLM-based models.
For each feature $f$, $A_f$ is the mean absolute deviation of model coefficients
from the human baseline; ``Share'' reports the fraction of total deviation
$\sum_f A_f$ accounted for by each feature.}
\label{tab:feature_misalignment}
\small
\begin{tabular}{l r r}
\toprule
Feature & Mean $A_f$ & Share (\%) \\
\midrule
Race: Pacific Islander   & 0.374 & 26.0 \\
Race: Native American    & 0.223 & 15.6 \\
Race: Black              & 0.194 & 13.5 \\
Spouse: Present          & 0.171 & 11.9 \\
Sex                      & 0.098 &  6.8 \\
Age                      & 0.095 &  6.6 \\
Earnings                 & 0.076 &  5.3 \\
Race: Asian              & 0.072 &  5.0 \\
Spouse: Unmarried        & 0.070 &  4.9 \\
Education                & 0.062 &  4.3 \\
\bottomrule
\end{tabular}
\end{table}

%% file: Tables/misalign_black_spouse_v2.tex
\begin{table*}[h]
\centering
\caption{Misalignment Analysis: Groups with Largest Deviation from Human Baseline}
\label{tab:worst_alignment}
\small
\setlength{\tabcolsep}{4pt}
\begin{tabular}{llc|ccccc}
\toprule
Activity & Group & Human & GPT-4o & Claude-3.7 & Deepseek-V3 & Llama-3.3 & Qwen-2.5 \\
\midrule
Work & Race: Black 
& \textbf{0.216}\tiny{$\pm$0.101} 
& \textbf{0.059}\tiny{$\pm$0.033} 
& \textbf{0.075}\tiny{$\pm$0.035} 
& \underline{-0.008}\tiny{$\pm$0.073} 
& \textbf{0.163}\tiny{$\pm$0.071} 
& 0.002\tiny{$\pm$0.043} \\
Work & Spouse: Present 
& \textbf{-0.236}\tiny{$\pm$0.065} 
& \underline{\textbf{0.046}}\tiny{$\pm$0.022} 
& \textbf{-0.119}\tiny{$\pm$0.023} 
& -0.030\tiny{$\pm$0.050} 
& 0.037\tiny{$\pm$0.046} 
& \textbf{-0.036}\tiny{$\pm$0.029} \\
\midrule
Leisure & Race: Black 
& \textbf{0.164}\tiny{$\pm$0.098} 
& \textbf{-0.053}\tiny{$\pm$0.036} 
& \textbf{-0.100}\tiny{$\pm$0.039} 
& \underline{\textbf{-0.119}}\tiny{$\pm$0.077} 
& -0.022\tiny{$\pm$0.077} 
& \textbf{-0.051}\tiny{$\pm$0.045} \\
Leisure & Spouse: Present 
& \textbf{-0.249}\tiny{$\pm$0.062} 
& \textbf{-0.180}\tiny{$\pm$0.023} 
& \textbf{-0.210}\tiny{$\pm$0.025} 
& \textbf{-0.106}\tiny{$\pm$0.052} 
& \underline{-0.017}\tiny{$\pm$0.049} 
& \textbf{-0.063}\tiny{$\pm$0.030} \\
\midrule
Sleep/PC & Race: Black 
& \textbf{0.189}\tiny{$\pm$0.072} 
& \textbf{0.040}\tiny{$\pm$0.031} 
& 0.001\tiny{$\pm$0.034} 
& -0.039\tiny{$\pm$0.072} 
& 0.019\tiny{$\pm$0.077} 
& \underline{-0.039}\tiny{$\pm$0.044} \\
Sleep/PC & Spouse: Present 
& \textbf{-0.187}\tiny{$\pm$0.044} 
& \underline{\textbf{0.049}}\tiny{$\pm$0.021} 
& \textbf{-0.105}\tiny{$\pm$0.023} 
& -0.023\tiny{$\pm$0.049} 
& -0.005\tiny{$\pm$0.050} 
& \textbf{-0.035}\tiny{$\pm$0.029} \\
\bottomrule
\multicolumn{8}{p{0.98\linewidth}}{\scriptsize 
\textit{Notes:} Sleep/PC denotes \emph{Sleep and Personal Care}. Entries are estimated coefficients from the constrained time-allocation model, reported as estimate $\pm$ 95\% confidence interval. 
Coefficients are interpreted \emph{relative to the reference activity} (``Other'') under the normalization $\theta_O=\mathbf{0}$: a positive coefficient means the subgroup tends to allocate \emph{more} time to the listed activity (vs.\ ``Other'') than the reference group; a negative coefficient means \emph{less} time. 
We compare coefficients across humans and models to diagnose which decision factors are over- or under-weighted by each LLM. 
\textbf{Bold} indicates statistical significance ($p<0.05$). 
\underline{Underline} marks, within each activity, the model with the largest absolute deviation from the human coefficient for that group (worst alignment).} \\
\end{tabular}
\vspace{-10pt}
\end{table*}

%% file: Tables/invariance.tex
\begin{table}[t]
\centering
\caption{Parameter drift under mild covariate shifts (lower is better). Bold indicates the smaller drift within each shift (\textsc{XChoice} vs.\ OLS).}
\label{tab:drift_summary}
\small
\sisetup{
  table-number-alignment = center,
  round-mode = places,
  round-precision = 4
}
\setlength{\tabcolsep}{6pt}
\renewcommand{\arraystretch}{1.15}

\resizebox{\columnwidth}{!}{%
\begin{tabular}{ll
                S[table-format=1.4]
                S[table-format=1.4]
                S[table-format=1.4]}
\toprule
\textbf{Shift} & \textbf{Method} & {\textbf{MAD}} & {\textbf{RelL2}} & {\textbf{1-CosSim}} \\
\midrule
\multirow{2}{*}{Earnings lift} &
\textsc{XChoice} & \textbf{0.0045} & \textbf{0.0251} & \textbf{0.0003} \\
& OLS      & 0.0380          & 0.0432          & 0.0008          \\
\addlinespace[2pt]
\multirow{2}{*}{Age band lift} &
\textsc{XChoice} & \textbf{0.0012} & \textbf{0.0055} & \textbf{0.0000} \\
& OLS      & 0.0169          & 0.0240          & 0.0003          \\
\addlinespace[2pt]
\multirow{2}{*}{Race mix shift} &
\textsc{XChoice} & \textbf{0.0374} & \textbf{0.1487} & \textbf{0.0111} \\
& OLS      & 0.4716          & 0.7522          & 0.1881          \\
\addlinespace[2pt]
\multirow{2}{*}{Spouse mix shift} &
\textsc{XChoice} & \textbf{0.0205} & \textbf{0.1088} & \textbf{0.0059} \\
& OLS      & 0.2190          & 0.3724          & 0.0692          \\
\bottomrule
\end{tabular}%
}
\vspace{-18pt}
\end{table}

%% file: Tables/RAG.tex
\begin{table}[t]
\centering
\small
\caption{RAG mitigation results. Cosine similarity to the human baseline.}
\label{tab:rag_cossim_results}
\resizebox{\columnwidth}{!}{%
\begin{tabular}{llccc}
\toprule
Model & Subgroup & Baseline & RAG& $\Delta$ \\
\midrule
GPT-4o      & Race: Black     & -0.827 & -0.782 & +0.045 \\
      & Spouse: Present & 0.276 & 0.741 & +0.465 \\
DeepSeek-V3 & Race: Black     & -0.774 & -0.673 & +0.101 \\
 & Spouse: Present & 0.704 & -0.015 & -0.719 \\
\bottomrule
\end{tabular}%
}
\end{table}

%% file: Tables/summary_stat.tex
\begin{table*}[h!]
\centering
\caption{Summary statistics for attributes and time-use variables (N = 4{,}307).}
\label{tab:summary_stat}
\small
\begin{tabular}{lrrrrrrr}
\toprule
Variable & Mean & Std.\ Dev. & Min & 25th pctl & Median & 75th pctl & Max \\
\midrule
\multicolumn{8}{l}{\textbf{Panel A: Continuous attributes (standardized)}}\\
Age                          &  0.000 & 1.000 & -2.054 & -0.827 & -0.077 &  0.809 &  2.718 \\
Education (\texttt{edu})     &  0.000 & 1.000 & -1.347 & -1.347 &  0.498 &  0.498 &  1.420 \\
Weekly earnings (\texttt{EARNWEEK}) &  0.000 & 1.000 & -1.621 & -0.765 & -0.195 &  0.637 &  1.945 \\
\midrule
\multicolumn{8}{l}{\textbf{Panel B: Binary and indicator attributes}}\\
Male                         &  0.507 & 0.500 & 0.000 & \multicolumn{3}{c}{--} & 1.000 \\
Spouse present               &  0.542 & 0.498 & 0.000 & \multicolumn{3}{c}{--} & 1.000 \\
Unmarried partner present    &  0.058 & 0.233 & 0.000 & \multicolumn{3}{c}{--} & 1.000 \\
Race: Black                  &  0.107 & 0.309 & 0.000 & \multicolumn{3}{c}{--} & 1.000 \\
Race: Native American        &  0.008 & 0.089 & 0.000 & \multicolumn{3}{c}{--} & 1.000 \\
Race: Asian                  &  0.068 & 0.252 & 0.000 & \multicolumn{3}{c}{--} & 1.000 \\
Race: Pacific Islander       &  0.002 & 0.043 & 0.000 & \multicolumn{3}{c}{--} & 1.000 \\
\midrule
\multicolumn{8}{l}{\textbf{Panel C: Time-use variables (minutes per day)}}\\
Work                         & 246.97 & 260.66 &   0.00 &   0.00 & 135.00 & 488.00 & 1390.00 \\
Leisure                      & 266.33 & 197.41 &   0.00 & 120.00 & 225.00 & 385.00 & 1237.00 \\
Sleep and personal care      & 578.16 & 136.48 &   0.00 & 500.00 & 570.00 & 645.00 & 1415.00 \\
Other                        & 348.53 & 210.83 &   0.00 & 185.00 & 305.00 & 480.00 & 1413.00 \\
\bottomrule
\end{tabular}

\vspace{0.5ex}
\footnotesize Note: Continuous predictors in Panel A are standardized to have mean zero and unit variance. 
For binary and indicator variables in Panel B, we report mean, standard deviation, minimum, and maximum; 
quantiles are not reported (indicated by ``--'').
\end{table*}

%% file: Tables/est_res.tex
\begin{table*}[h]
\centering
\caption{Estimated Coefficients by Activity and Model (Human Baseline vs. LLMs)}
\label{tab:coef_appendix}
\small
\setlength{\tabcolsep}{4pt}
\begin{tabular}{lc|ccccc}
\toprule
Feature & Human & GPT-4o & Claude-3.7 & Deepseek-V3 & Llama-3.3 & Qwen-2.5 \\
\midrule
\multicolumn{7}{c}{\textbf{Activity: Leisure}} \\
\midrule
Constant & \textbf{0.326} $\pm$ \textcolor{gray}{\tiny{0.102}} & \textbf{1.035} $\pm$ \textcolor{gray}{\tiny{0.041}} & \textbf{1.344} $\pm$ \textcolor{gray}{\tiny{0.043}} & \textbf{1.704} $\pm$ \textcolor{gray}{\tiny{0.090}} & \textbf{1.331} $\pm$ \textcolor{gray}{\tiny{0.088}} & \textbf{1.752} $\pm$ \textcolor{gray}{\tiny{0.051}} \\
Age & \textbf{0.055} $\pm$ \textcolor{gray}{\tiny{0.029}} & \textbf{-0.146} $\pm$ \textcolor{gray}{\tiny{0.011}} & \textbf{0.102} $\pm$ \textcolor{gray}{\tiny{0.012}} & \textbf{0.050} $\pm$ \textcolor{gray}{\tiny{0.024}} & \textbf{0.066} $\pm$ \textcolor{gray}{\tiny{0.023}} & \textbf{0.038} $\pm$ \textcolor{gray}{\tiny{0.014}} \\
Sex & \textbf{-0.324} $\pm$ \textcolor{gray}{\tiny{0.059}} & \textbf{-0.157} $\pm$ \textcolor{gray}{\tiny{0.023}} & \textbf{-0.225} $\pm$ \textcolor{gray}{\tiny{0.024}} & \textbf{-0.186} $\pm$ \textcolor{gray}{\tiny{0.050}} & \textbf{-0.168} $\pm$ \textcolor{gray}{\tiny{0.049}} & \textbf{-0.141} $\pm$ \textcolor{gray}{\tiny{0.028}} \\
Education & -0.025 $\pm$ \textcolor{gray}{\tiny{0.032}} & \textbf{-0.014} $\pm$ \textcolor{gray}{\tiny{0.012}} & \textbf{-0.044} $\pm$ \textcolor{gray}{\tiny{0.013}} & \textbf{0.044} $\pm$ \textcolor{gray}{\tiny{0.027}} & \textbf{0.059} $\pm$ \textcolor{gray}{\tiny{0.026}} & 0.012 $\pm$ \textcolor{gray}{\tiny{0.016}} \\
Earnings & \textbf{-0.060} $\pm$ \textcolor{gray}{\tiny{0.033}} & \textbf{0.069} $\pm$ \textcolor{gray}{\tiny{0.013}} & \textbf{0.045} $\pm$ \textcolor{gray}{\tiny{0.014}} & -0.002 $\pm$ \textcolor{gray}{\tiny{0.029}} & \textbf{0.040} $\pm$ \textcolor{gray}{\tiny{0.029}} & \textbf{-0.041} $\pm$ \textcolor{gray}{\tiny{0.016}} \\
Asian & -0.072 $\pm$ \textcolor{gray}{\tiny{0.118}} & \textbf{-0.132} $\pm$ \textcolor{gray}{\tiny{0.048}} & \textbf{-0.103} $\pm$ \textcolor{gray}{\tiny{0.048}} & -0.075 $\pm$ \textcolor{gray}{\tiny{0.100}} & \textbf{0.125} $\pm$ \textcolor{gray}{\tiny{0.122}} & -0.052 $\pm$ \textcolor{gray}{\tiny{0.055}} \\
Black & \textbf{0.164} $\pm$ \textcolor{gray}{\tiny{0.098}} & \textbf{-0.053} $\pm$ \textcolor{gray}{\tiny{0.036}} & \textbf{-0.100} $\pm$ \textcolor{gray}{\tiny{0.039}} & \textbf{-0.119} $\pm$ \textcolor{gray}{\tiny{0.077}} & -0.022 $\pm$ \textcolor{gray}{\tiny{0.077}} & \textbf{-0.051} $\pm$ \textcolor{gray}{\tiny{0.045}} \\
Native Am. & 0.027 $\pm$ \textcolor{gray}{\tiny{0.321}} & 0.088 $\pm$ \textcolor{gray}{\tiny{0.139}} & -0.038 $\pm$ \textcolor{gray}{\tiny{0.138}} & -0.154 $\pm$ \textcolor{gray}{\tiny{0.261}} & \textbf{-0.299} $\pm$ \textcolor{gray}{\tiny{0.216}} & \textbf{-0.574} $\pm$ \textcolor{gray}{\tiny{0.108}} \\
Pac. Isl. & 0.130 $\pm$ \textcolor{gray}{\tiny{0.795}} & 0.026 $\pm$ \textcolor{gray}{\tiny{0.282}} & -0.005 $\pm$ \textcolor{gray}{\tiny{0.297}} & -0.067 $\pm$ \textcolor{gray}{\tiny{0.517}} & -0.164 $\pm$ \textcolor{gray}{\tiny{0.470}} & \textbf{-0.390} $\pm$ \textcolor{gray}{\tiny{0.265}} \\
Married & \textbf{-0.249} $\pm$ \textcolor{gray}{\tiny{0.062}} & \textbf{-0.180} $\pm$ \textcolor{gray}{\tiny{0.023}} & \textbf{-0.210} $\pm$ \textcolor{gray}{\tiny{0.025}} & \textbf{-0.106} $\pm$ \textcolor{gray}{\tiny{0.052}} & -0.017 $\pm$ \textcolor{gray}{\tiny{0.049}} & \textbf{-0.063} $\pm$ \textcolor{gray}{\tiny{0.030}} \\
Unmarried & 0.036 $\pm$ \textcolor{gray}{\tiny{0.131}} & \textbf{-0.141} $\pm$ \textcolor{gray}{\tiny{0.055}} & \textbf{-0.122} $\pm$ \textcolor{gray}{\tiny{0.054}} & 0.036 $\pm$ \textcolor{gray}{\tiny{0.113}} & -0.030 $\pm$ \textcolor{gray}{\tiny{0.105}} & -0.051 $\pm$ \textcolor{gray}{\tiny{0.061}} \\
\midrule
\multicolumn{7}{c}{\textbf{Activity: Sleep and Personal Care}} \\
\midrule
Constant & \textbf{0.677} $\pm$ \textcolor{gray}{\tiny{0.074}} & \textbf{1.537} $\pm$ \textcolor{gray}{\tiny{0.036}} & \textbf{1.942} $\pm$ \textcolor{gray}{\tiny{0.039}} & \textbf{2.273} $\pm$ \textcolor{gray}{\tiny{0.085}} & \textbf{1.112} $\pm$ \textcolor{gray}{\tiny{0.090}} & \textbf{1.912} $\pm$ \textcolor{gray}{\tiny{0.050}} \\
Age & -0.011 $\pm$ \textcolor{gray}{\tiny{0.021}} & \textbf{-0.226} $\pm$ \textcolor{gray}{\tiny{0.010}} & \textbf{-0.014} $\pm$ \textcolor{gray}{\tiny{0.011}} & -0.009 $\pm$ \textcolor{gray}{\tiny{0.023}} & 0.005 $\pm$ \textcolor{gray}{\tiny{0.023}} & \textbf{0.019} $\pm$ \textcolor{gray}{\tiny{0.014}} \\
Sex & \textbf{-0.064} $\pm$ \textcolor{gray}{\tiny{0.042}} & \textbf{-0.089} $\pm$ \textcolor{gray}{\tiny{0.020}} & \textbf{-0.130} $\pm$ \textcolor{gray}{\tiny{0.022}} & \textbf{-0.127} $\pm$ \textcolor{gray}{\tiny{0.048}} & \textbf{-0.076} $\pm$ \textcolor{gray}{\tiny{0.050}} & \textbf{-0.116} $\pm$ \textcolor{gray}{\tiny{0.028}} \\
Education & \textbf{-0.038} $\pm$ \textcolor{gray}{\tiny{0.023}} & \textbf{-0.021} $\pm$ \textcolor{gray}{\tiny{0.011}} & \textbf{-0.053} $\pm$ \textcolor{gray}{\tiny{0.012}} & \textbf{0.049} $\pm$ \textcolor{gray}{\tiny{0.026}} & \textbf{0.037} $\pm$ \textcolor{gray}{\tiny{0.027}} & 0.006 $\pm$ \textcolor{gray}{\tiny{0.015}} \\
Earnings & -0.018 $\pm$ \textcolor{gray}{\tiny{0.023}} & \textbf{0.108} $\pm$ \textcolor{gray}{\tiny{0.012}} & \textbf{0.091} $\pm$ \textcolor{gray}{\tiny{0.013}} & 0.008 $\pm$ \textcolor{gray}{\tiny{0.027}} & 0.028 $\pm$ \textcolor{gray}{\tiny{0.030}} & \textbf{-0.059} $\pm$ \textcolor{gray}{\tiny{0.016}} \\
Asian & 0.003 $\pm$ \textcolor{gray}{\tiny{0.078}} & \textbf{-0.052} $\pm$ \textcolor{gray}{\tiny{0.041}} & -0.019 $\pm$ \textcolor{gray}{\tiny{0.042}} & -0.036 $\pm$ \textcolor{gray}{\tiny{0.094}} & 0.081 $\pm$ \textcolor{gray}{\tiny{0.124}} & -0.021 $\pm$ \textcolor{gray}{\tiny{0.054}} \\
Black & \textbf{0.189} $\pm$ \textcolor{gray}{\tiny{0.072}} & \textbf{0.040} $\pm$ \textcolor{gray}{\tiny{0.031}} & 0.001 $\pm$ \textcolor{gray}{\tiny{0.034}} & -0.039 $\pm$ \textcolor{gray}{\tiny{0.072}} & 0.019 $\pm$ \textcolor{gray}{\tiny{0.077}} & -0.039 $\pm$ \textcolor{gray}{\tiny{0.044}} \\
Native Am. & 0.033 $\pm$ \textcolor{gray}{\tiny{0.228}} & 0.089 $\pm$ \textcolor{gray}{\tiny{0.123}} & 0.031 $\pm$ \textcolor{gray}{\tiny{0.123}} & -0.082 $\pm$ \textcolor{gray}{\tiny{0.243}} & \textbf{-0.247} $\pm$ \textcolor{gray}{\tiny{0.217}} & \textbf{-0.531} $\pm$ \textcolor{gray}{\tiny{0.103}} \\
Pac. Isl. & 0.433 $\pm$ \textcolor{gray}{\tiny{0.559}} & 0.052 $\pm$ \textcolor{gray}{\tiny{0.251}} & 0.074 $\pm$ \textcolor{gray}{\tiny{0.266}} & -0.031 $\pm$ \textcolor{gray}{\tiny{0.486}} & -0.081 $\pm$ \textcolor{gray}{\tiny{0.468}} & -0.219 $\pm$ \textcolor{gray}{\tiny{0.252}} \\
Married & \textbf{-0.187} $\pm$ \textcolor{gray}{\tiny{0.044}} & \textbf{0.049} $\pm$ \textcolor{gray}{\tiny{0.021}} & \textbf{-0.105} $\pm$ \textcolor{gray}{\tiny{0.023}} & -0.023 $\pm$ \textcolor{gray}{\tiny{0.049}} & -0.005 $\pm$ \textcolor{gray}{\tiny{0.050}} & \textbf{-0.035} $\pm$ \textcolor{gray}{\tiny{0.029}} \\
Unmarried & 0.008 $\pm$ \textcolor{gray}{\tiny{0.096}} & 0.006 $\pm$ \textcolor{gray}{\tiny{0.048}} & \textbf{-0.071} $\pm$ \textcolor{gray}{\tiny{0.048}} & -0.014 $\pm$ \textcolor{gray}{\tiny{0.107}} & -0.041 $\pm$ \textcolor{gray}{\tiny{0.106}} & -0.051 $\pm$ \textcolor{gray}{\tiny{0.060}} \\
\midrule
\multicolumn{7}{c}{\textbf{Activity: Work}} \\
\midrule
Constant & 0.076 $\pm$ \textcolor{gray}{\tiny{0.108}} & \textbf{1.265} $\pm$ \textcolor{gray}{\tiny{0.038}} & \textbf{1.741} $\pm$ \textcolor{gray}{\tiny{0.040}} & \textbf{2.127} $\pm$ \textcolor{gray}{\tiny{0.086}} & \textbf{2.152} $\pm$ \textcolor{gray}{\tiny{0.082}} & \textbf{2.043} $\pm$ \textcolor{gray}{\tiny{0.049}} \\
Age & 0.009 $\pm$ \textcolor{gray}{\tiny{0.031}} & \textbf{-0.353} $\pm$ \textcolor{gray}{\tiny{0.010}} & \textbf{-0.104} $\pm$ \textcolor{gray}{\tiny{0.011}} & \textbf{-0.115} $\pm$ \textcolor{gray}{\tiny{0.023}} & \textbf{-0.169} $\pm$ \textcolor{gray}{\tiny{0.021}} & \textbf{-0.092} $\pm$ \textcolor{gray}{\tiny{0.014}} \\
Sex & \textbf{-0.221} $\pm$ \textcolor{gray}{\tiny{0.063}} & \textbf{-0.109} $\pm$ \textcolor{gray}{\tiny{0.021}} & \textbf{-0.164} $\pm$ \textcolor{gray}{\tiny{0.022}} & \textbf{-0.146} $\pm$ \textcolor{gray}{\tiny{0.048}} & \textbf{-0.384} $\pm$ \textcolor{gray}{\tiny{0.046}} & \textbf{-0.116} $\pm$ \textcolor{gray}{\tiny{0.028}} \\
Education & \textbf{-0.099} $\pm$ \textcolor{gray}{\tiny{0.034}} & \textbf{-0.029} $\pm$ \textcolor{gray}{\tiny{0.011}} & \textbf{-0.081} $\pm$ \textcolor{gray}{\tiny{0.012}} & \textbf{0.033} $\pm$ \textcolor{gray}{\tiny{0.026}} & \textbf{0.072} $\pm$ \textcolor{gray}{\tiny{0.025}} & -0.011 $\pm$ \textcolor{gray}{\tiny{0.015}} \\
Earnings & \textbf{0.112} $\pm$ \textcolor{gray}{\tiny{0.034}} & \textbf{0.184} $\pm$ \textcolor{gray}{\tiny{0.012}} & \textbf{0.175} $\pm$ \textcolor{gray}{\tiny{0.013}} & \textbf{0.068} $\pm$ \textcolor{gray}{\tiny{0.028}} & \textbf{0.234} $\pm$ \textcolor{gray}{\tiny{0.028}} & \textbf{0.037} $\pm$ \textcolor{gray}{\tiny{0.015}} \\
Asian & -0.045 $\pm$ \textcolor{gray}{\tiny{0.120}} & \textbf{-0.074} $\pm$ \textcolor{gray}{\tiny{0.042}} & 0.016 $\pm$ \textcolor{gray}{\tiny{0.043}} & -0.013 $\pm$ \textcolor{gray}{\tiny{0.094}} & \textbf{0.371} $\pm$ \textcolor{gray}{\tiny{0.113}} & -0.038 $\pm$ \textcolor{gray}{\tiny{0.053}} \\
Black & \textbf{0.216} $\pm$ \textcolor{gray}{\tiny{0.101}} & \textbf{0.059} $\pm$ \textcolor{gray}{\tiny{0.033}} & \textbf{0.075} $\pm$ \textcolor{gray}{\tiny{0.035}} & -0.008 $\pm$ \textcolor{gray}{\tiny{0.073}} & \textbf{0.163} $\pm$ \textcolor{gray}{\tiny{0.071}} & 0.002 $\pm$ \textcolor{gray}{\tiny{0.043}} \\
Native Am. & 0.055 $\pm$ \textcolor{gray}{\tiny{0.336}} & \textbf{0.131} $\pm$ \textcolor{gray}{\tiny{0.127}} & 0.095 $\pm$ \textcolor{gray}{\tiny{0.125}} & -0.051 $\pm$ \textcolor{gray}{\tiny{0.245}} & \textbf{-0.331} $\pm$ \textcolor{gray}{\tiny{0.197}} & \textbf{-0.436} $\pm$ \textcolor{gray}{\tiny{0.100}} \\
Pac. Isl. & 0.319 $\pm$ \textcolor{gray}{\tiny{0.769}} & 0.139 $\pm$ \textcolor{gray}{\tiny{0.257}} & 0.120 $\pm$ \textcolor{gray}{\tiny{0.271}} & -0.213 $\pm$ \textcolor{gray}{\tiny{0.497}} & -0.185 $\pm$ \textcolor{gray}{\tiny{0.434}} & \textbf{-0.256} $\pm$ \textcolor{gray}{\tiny{0.250}} \\
Married & \textbf{-0.236} $\pm$ \textcolor{gray}{\tiny{0.065}} & \textbf{0.046} $\pm$ \textcolor{gray}{\tiny{0.022}} & \textbf{-0.119} $\pm$ \textcolor{gray}{\tiny{0.023}} & -0.030 $\pm$ \textcolor{gray}{\tiny{0.050}} & 0.037 $\pm$ \textcolor{gray}{\tiny{0.046}} & \textbf{-0.036} $\pm$ \textcolor{gray}{\tiny{0.029}} \\
Unmarried & 0.031 $\pm$ \textcolor{gray}{\tiny{0.135}} & -0.011 $\pm$ \textcolor{gray}{\tiny{0.050}} & \textbf{-0.095} $\pm$ \textcolor{gray}{\tiny{0.049}} & 0.006 $\pm$ \textcolor{gray}{\tiny{0.108}} & -0.039 $\pm$ \textcolor{gray}{\tiny{0.096}} & -0.056 $\pm$ \textcolor{gray}{\tiny{0.059}} \\
\bottomrule
\multicolumn{7}{l}{\scriptsize \textit{Note:} Coefficients significant at $p<0.05$ are \textbf{bolded}. Margins denote 95\% CI.} \\
\end{tabular}
\end{table*}

%% file: Tables/ols_res.tex
\begin{table*}[h]
\centering
\caption{Estimated Coefficients by Activity and Model (Human Baseline vs. LLMs). Values represent OLS coefficients $\pm$ 95\% CI margin.}
\label{tab:ols_coef_appendix}
\small
\setlength{\tabcolsep}{3.5pt}
\begin{tabular}{lc|ccccc}
\toprule
Feature & Human & GPT-4o & Claude & DeepSeek & Llama & Qwen \\
\midrule
\multicolumn{7}{c}{\textbf{Activity: Leisure}} \\
\midrule
Age & \textbf{0.061} $\pm$ \textcolor{gray}{\tiny{0.028}} & \textbf{0.115} $\pm$ \textcolor{gray}{\tiny{0.006}} & \textbf{0.151} $\pm$ \textcolor{gray}{\tiny{0.006}} & \textbf{0.110} $\pm$ \textcolor{gray}{\tiny{0.008}} & \textbf{0.190} $\pm$ \textcolor{gray}{\tiny{0.010}} & \textbf{0.096} $\pm$ \textcolor{gray}{\tiny{0.005}} \\
Sex & \textbf{-3.508} $\pm$ \textcolor{gray}{\tiny{0.841}} & \textbf{-0.894} $\pm$ \textcolor{gray}{\tiny{0.185}} & \textbf{-1.280} $\pm$ \textcolor{gray}{\tiny{0.164}} & \textbf{-0.805} $\pm$ \textcolor{gray}{\tiny{0.248}} & \textbf{1.483} $\pm$ \textcolor{gray}{\tiny{0.289}} & \textbf{-0.458} $\pm$ \textcolor{gray}{\tiny{0.135}} \\
Education & 0.171 $\pm$ \textcolor{gray}{\tiny{0.425}} & \textbf{0.131} $\pm$ \textcolor{gray}{\tiny{0.094}} & \textbf{0.164} $\pm$ \textcolor{gray}{\tiny{0.083}} & 0.048 $\pm$ \textcolor{gray}{\tiny{0.125}} & 0.088 $\pm$ \textcolor{gray}{\tiny{0.146}} & \textbf{0.255} $\pm$ \textcolor{gray}{\tiny{0.068}} \\
Earnings & \textbf{-0.001} $\pm$ \textcolor{gray}{\tiny{0.001}} & \textbf{-0.001} $\pm$ \textcolor{gray}{\tiny{0.000}} & \textbf{-0.001} $\pm$ \textcolor{gray}{\tiny{0.000}} & \textbf{-0.001} $\pm$ \textcolor{gray}{\tiny{0.000}} & \textbf{-0.002} $\pm$ \textcolor{gray}{\tiny{0.000}} & \textbf{-0.001} $\pm$ \textcolor{gray}{\tiny{0.000}} \\
Asian & -0.950 $\pm$ \textcolor{gray}{\tiny{1.641}} & \textbf{-1.079} $\pm$ \textcolor{gray}{\tiny{0.362}} & \textbf{-1.376} $\pm$ \textcolor{gray}{\tiny{0.321}} & \textbf{-0.783} $\pm$ \textcolor{gray}{\tiny{0.483}} & \textbf{-2.420} $\pm$ \textcolor{gray}{\tiny{0.565}} & \textbf{-0.465} $\pm$ \textcolor{gray}{\tiny{0.264}} \\
Black & 0.465 $\pm$ \textcolor{gray}{\tiny{1.335}} & \textbf{-1.509} $\pm$ \textcolor{gray}{\tiny{0.294}} & \textbf{-1.855} $\pm$ \textcolor{gray}{\tiny{0.261}} & \textbf{-1.566} $\pm$ \textcolor{gray}{\tiny{0.393}} & \textbf{-2.150} $\pm$ \textcolor{gray}{\tiny{0.459}} & \textbf{-0.658} $\pm$ \textcolor{gray}{\tiny{0.214}} \\
Native Am. & 0.135 $\pm$ \textcolor{gray}{\tiny{4.581}} & -0.134 $\pm$ \textcolor{gray}{\tiny{1.009}} & \textbf{-1.244} $\pm$ \textcolor{gray}{\tiny{0.896}} & \textbf{-1.355} $\pm$ \textcolor{gray}{\tiny{1.348}} & -0.443 $\pm$ \textcolor{gray}{\tiny{1.575}} & \textbf{-2.646} $\pm$ \textcolor{gray}{\tiny{0.734}} \\
Pac. Isl. & -2.182 $\pm$ \textcolor{gray}{\tiny{9.411}} & -0.694 $\pm$ \textcolor{gray}{\tiny{2.073}} & -1.283 $\pm$ \textcolor{gray}{\tiny{1.840}} & 0.400 $\pm$ \textcolor{gray}{\tiny{2.769}} & -0.684 $\pm$ \textcolor{gray}{\tiny{3.236}} & \textbf{-3.225} $\pm$ \textcolor{gray}{\tiny{1.507}} \\
Married & \textbf{-1.600} $\pm$ \textcolor{gray}{\tiny{0.879}} & \textbf{-3.515} $\pm$ \textcolor{gray}{\tiny{0.194}} & \textbf{-1.669} $\pm$ \textcolor{gray}{\tiny{0.172}} & \textbf{-1.327} $\pm$ \textcolor{gray}{\tiny{0.259}} & \textbf{-0.793} $\pm$ \textcolor{gray}{\tiny{0.302}} & \textbf{-0.500} $\pm$ \textcolor{gray}{\tiny{0.141}} \\
Unmarried & 0.606 $\pm$ \textcolor{gray}{\tiny{1.818}} & \textbf{-2.175} $\pm$ \textcolor{gray}{\tiny{0.401}} & \textbf{-0.712} $\pm$ \textcolor{gray}{\tiny{0.356}} & \textbf{0.685} $\pm$ \textcolor{gray}{\tiny{0.537}} & 0.042 $\pm$ \textcolor{gray}{\tiny{0.626}} & 0.062 $\pm$ \textcolor{gray}{\tiny{0.292}} \\
\midrule
\multicolumn{7}{c}{\textbf{Activity: Sleep and Personal Care}} \\
\midrule
Age & \textbf{-0.049} $\pm$ \textcolor{gray}{\tiny{0.019}} & \textbf{0.010} $\pm$ \textcolor{gray}{\tiny{0.004}} & \textbf{0.025} $\pm$ \textcolor{gray}{\tiny{0.003}} & \textbf{0.067} $\pm$ \textcolor{gray}{\tiny{0.008}} & \textbf{0.101} $\pm$ \textcolor{gray}{\tiny{0.011}} & \textbf{0.079} $\pm$ \textcolor{gray}{\tiny{0.005}} \\
Sex & \textbf{2.001} $\pm$ \textcolor{gray}{\tiny{0.575}} & \textbf{0.599} $\pm$ \textcolor{gray}{\tiny{0.112}} & \textbf{0.471} $\pm$ \textcolor{gray}{\tiny{0.084}} & \textbf{0.652} $\pm$ \textcolor{gray}{\tiny{0.235}} & \textbf{3.415} $\pm$ \textcolor{gray}{\tiny{0.319}} & 0.138 $\pm$ \textcolor{gray}{\tiny{0.163}} \\
Education & -0.023 $\pm$ \textcolor{gray}{\tiny{0.291}} & -0.008 $\pm$ \textcolor{gray}{\tiny{0.057}} & \textbf{0.166} $\pm$ \textcolor{gray}{\tiny{0.042}} & \textbf{0.301} $\pm$ \textcolor{gray}{\tiny{0.119}} & \textbf{-0.364} $\pm$ \textcolor{gray}{\tiny{0.162}} & \textbf{0.139} $\pm$ \textcolor{gray}{\tiny{0.082}} \\
Earnings & \textbf{-0.001} $\pm$ \textcolor{gray}{\tiny{0.000}} & \textbf{-0.000} $\pm$ \textcolor{gray}{\tiny{0.000}} & \textbf{-0.001} $\pm$ \textcolor{gray}{\tiny{0.000}} & \textbf{-0.001} $\pm$ \textcolor{gray}{\tiny{0.000}} & \textbf{-0.003} $\pm$ \textcolor{gray}{\tiny{0.000}} & \textbf{-0.002} $\pm$ \textcolor{gray}{\tiny{0.000}} \\
Asian & 0.863 $\pm$ \textcolor{gray}{\tiny{1.124}} & \textbf{0.654} $\pm$ \textcolor{gray}{\tiny{0.219}} & -0.034 $\pm$ \textcolor{gray}{\tiny{0.163}} & -0.100 $\pm$ \textcolor{gray}{\tiny{0.458}} & \textbf{-3.022} $\pm$ \textcolor{gray}{\tiny{0.624}} & \textbf{0.419} $\pm$ \textcolor{gray}{\tiny{0.319}} \\
Black & \textbf{1.604} $\pm$ \textcolor{gray}{\tiny{0.914}} & \textbf{0.681} $\pm$ \textcolor{gray}{\tiny{0.178}} & \textbf{-0.336} $\pm$ \textcolor{gray}{\tiny{0.133}} & 0.144 $\pm$ \textcolor{gray}{\tiny{0.373}} & \textbf{-1.147} $\pm$ \textcolor{gray}{\tiny{0.507}} & \textbf{-0.457} $\pm$ \textcolor{gray}{\tiny{0.258}} \\
Native Am. & 0.244 $\pm$ \textcolor{gray}{\tiny{3.136}} & -0.178 $\pm$ \textcolor{gray}{\tiny{0.611}} & -0.426 $\pm$ \textcolor{gray}{\tiny{0.456}} & -0.065 $\pm$ \textcolor{gray}{\tiny{1.277}} & 0.738 $\pm$ \textcolor{gray}{\tiny{1.740}} & \textbf{-1.945} $\pm$ \textcolor{gray}{\tiny{0.887}} \\
Pac. Isl. & \textbf{7.311} $\pm$ \textcolor{gray}{\tiny{6.442}} & -0.536 $\pm$ \textcolor{gray}{\tiny{1.255}} & 0.047 $\pm$ \textcolor{gray}{\tiny{0.936}} & 2.563 $\pm$ \textcolor{gray}{\tiny{2.624}} & 1.467 $\pm$ \textcolor{gray}{\tiny{3.574}} & 1.180 $\pm$ \textcolor{gray}{\tiny{1.822}} \\
Married & \textbf{-1.289} $\pm$ \textcolor{gray}{\tiny{0.602}} & \textbf{2.234} $\pm$ \textcolor{gray}{\tiny{0.117}} & \textbf{0.615} $\pm$ \textcolor{gray}{\tiny{0.087}} & \textbf{0.828} $\pm$ \textcolor{gray}{\tiny{0.245}} & -0.282 $\pm$ \textcolor{gray}{\tiny{0.334}} & \textbf{0.281} $\pm$ \textcolor{gray}{\tiny{0.170}} \\
Unmarried & -0.563 $\pm$ \textcolor{gray}{\tiny{1.244}} & \textbf{1.544} $\pm$ \textcolor{gray}{\tiny{0.243}} & \textbf{0.435} $\pm$ \textcolor{gray}{\tiny{0.181}} & \textbf{-0.621} $\pm$ \textcolor{gray}{\tiny{0.509}} & -0.062 $\pm$ \textcolor{gray}{\tiny{0.692}} & 0.055 $\pm$ \textcolor{gray}{\tiny{0.353}} \\
\midrule
\multicolumn{7}{c}{\textbf{Activity: Work}} \\
\midrule
Age & 0.000 $\pm$ \textcolor{gray}{\tiny{0.038}} & \textbf{-0.272} $\pm$ \textcolor{gray}{\tiny{0.011}} & \textbf{-0.191} $\pm$ \textcolor{gray}{\tiny{0.009}} & \textbf{-0.200} $\pm$ \textcolor{gray}{\tiny{0.014}} & \textbf{-0.329} $\pm$ \textcolor{gray}{\tiny{0.023}} & \textbf{-0.194} $\pm$ \textcolor{gray}{\tiny{0.008}} \\
Sex & \textbf{-1.996} $\pm$ \textcolor{gray}{\tiny{1.115}} & -0.147 $\pm$ \textcolor{gray}{\tiny{0.316}} & \textbf{-0.591} $\pm$ \textcolor{gray}{\tiny{0.278}} & -0.166 $\pm$ \textcolor{gray}{\tiny{0.402}} & \textbf{-6.728} $\pm$ \textcolor{gray}{\tiny{0.683}} & 0.127 $\pm$ \textcolor{gray}{\tiny{0.246}} \\
Education & \textbf{-0.939} $\pm$ \textcolor{gray}{\tiny{0.564}} & \textbf{-0.288} $\pm$ \textcolor{gray}{\tiny{0.160}} & \textbf{-0.769} $\pm$ \textcolor{gray}{\tiny{0.141}} & \textbf{-0.304} $\pm$ \textcolor{gray}{\tiny{0.203}} & \textbf{0.622} $\pm$ \textcolor{gray}{\tiny{0.346}} & \textbf{-0.438} $\pm$ \textcolor{gray}{\tiny{0.125}} \\
Earnings & \textbf{0.002} $\pm$ \textcolor{gray}{\tiny{0.001}} & \textbf{0.003} $\pm$ \textcolor{gray}{\tiny{0.000}} & \textbf{0.003} $\pm$ \textcolor{gray}{\tiny{0.000}} & \textbf{0.002} $\pm$ \textcolor{gray}{\tiny{0.000}} & \textbf{0.006} $\pm$ \textcolor{gray}{\tiny{0.000}} & \textbf{0.002} $\pm$ \textcolor{gray}{\tiny{0.000}} \\
Asian & -0.430 $\pm$ \textcolor{gray}{\tiny{2.177}} & -0.144 $\pm$ \textcolor{gray}{\tiny{0.618}} & \textbf{1.144} $\pm$ \textcolor{gray}{\tiny{0.543}} & 0.768 $\pm$ \textcolor{gray}{\tiny{0.784}} & \textbf{6.669} $\pm$ \textcolor{gray}{\tiny{1.335}} & -0.156 $\pm$ \textcolor{gray}{\tiny{0.482}} \\
Black & 1.285 $\pm$ \textcolor{gray}{\tiny{1.770}} & \textbf{1.145} $\pm$ \textcolor{gray}{\tiny{0.502}} & \textbf{2.191} $\pm$ \textcolor{gray}{\tiny{0.442}} & \textbf{1.184} $\pm$ \textcolor{gray}{\tiny{0.638}} & \textbf{3.969} $\pm$ \textcolor{gray}{\tiny{1.084}} & \textbf{1.051} $\pm$ \textcolor{gray}{\tiny{0.391}} \\
Native Am. & 0.387 $\pm$ \textcolor{gray}{\tiny{6.076}} & 1.333 $\pm$ \textcolor{gray}{\tiny{1.724}} & \textbf{1.896} $\pm$ \textcolor{gray}{\tiny{1.516}} & 1.088 $\pm$ \textcolor{gray}{\tiny{2.187}} & -2.419 $\pm$ \textcolor{gray}{\tiny{3.720}} & 1.265 $\pm$ \textcolor{gray}{\tiny{1.341}} \\
Pac. Isl. & 0.798 $\pm$ \textcolor{gray}{\tiny{12.482}} & 2.287 $\pm$ \textcolor{gray}{\tiny{3.541}} & 1.595 $\pm$ \textcolor{gray}{\tiny{3.114}} & -4.069 $\pm$ \textcolor{gray}{\tiny{4.493}} & -1.736 $\pm$ \textcolor{gray}{\tiny{7.642}} & 0.058 $\pm$ \textcolor{gray}{\tiny{2.755}} \\
Married & \textbf{-1.375} $\pm$ \textcolor{gray}{\tiny{1.166}} & \textbf{1.525} $\pm$ \textcolor{gray}{\tiny{0.331}} & 0.020 $\pm$ \textcolor{gray}{\tiny{0.291}} & 0.359 $\pm$ \textcolor{gray}{\tiny{0.420}} & \textbf{1.001} $\pm$ \textcolor{gray}{\tiny{0.714}} & 0.197 $\pm$ \textcolor{gray}{\tiny{0.257}} \\
Unmarried & 0.117 $\pm$ \textcolor{gray}{\tiny{2.411}} & 0.563 $\pm$ \textcolor{gray}{\tiny{0.684}} & -0.518 $\pm$ \textcolor{gray}{\tiny{0.602}} & 0.065 $\pm$ \textcolor{gray}{\tiny{0.871}} & -0.293 $\pm$ \textcolor{gray}{\tiny{1.479}} & -0.226 $\pm$ \textcolor{gray}{\tiny{0.533}} \\
\bottomrule
\end{tabular}
\end{table*}